\documentclass[journal]{IEEEtran}
\usepackage[pagebackref,breaklinks,colorlinks]{hyperref}
\usepackage{url}
\usepackage[utf8]{inputenc} 
\usepackage[T1]{fontenc}    
\usepackage{hyperref}       
\usepackage{url}            
\usepackage{booktabs}       
\usepackage{amsfonts}       
\usepackage{nicefrac}       
\usepackage{microtype}      
\usepackage{xcolor}         
\usepackage{amsmath}
\usepackage{multicol}
\usepackage{multirow}
\usepackage{graphicx}
\usepackage{wrapfig}

\usepackage{textcomp}
\usepackage{xcolor}
\usepackage{bm}
\usepackage{bbm}
\usepackage[varbb]{newpxmath}
\usepackage{algorithm}
\usepackage{algorithmic} 
\usepackage{multirow}
\usepackage{caption}
\usepackage{subcaption}
\usepackage{supertabular}

\begin{document}


\title{Logit Margin Matters: Improving Transferable Targeted Adversarial Attack by Logit Calibration}

\author{
        Juanjuan Weng,
        Zhiming Luo,
        Zhun Zhong,
        Shaozi Li,
        Nicu Sebe
\IEEEcompsocitemizethanks{
\IEEEcompsocthanksitem  J. Weng, Z. Luo (Corresponding author), and S. Li are with the Department of Artificial Intelligence, Xiamen University,
Xiamen 361005, China.
\IEEEcompsocthanksitem Z. Zhong and N. Sebe are with the Department of Information Engineering and Computer Science, University of Trento, Trento 38100, Italy.
}
}

\maketitle
\begin{abstract}
Previous works have extensively studied the transferability of adversarial samples in untargeted black-box scenarios. However, it still remains challenging to craft targeted adversarial examples with higher transferability than non-targeted ones. Recent studies reveal that the traditional Cross-Entropy (CE) loss function is insufficient to learn transferable targeted adversarial examples due to the issue of vanishing gradient. In this work, we provide a comprehensive investigation of the CE loss function and find that the logit margin between the targeted and untargeted classes will quickly obtain saturation in CE, which largely limits the transferability. 
Therefore, in this paper, we devote to the goal of continually increasing the logit margin along the optimization to deal with the saturation
issue and propose two simple and effective logit calibration methods, which are achieved by downscaling the logits with a temperature factor and an adaptive margin, respectively.
Both of them can effectively encourage optimization to produce a larger logit margin and lead to higher transferability. Besides, we show that minimizing the cosine distance between the adversarial examples and the classifier weights of the target class can further improve the transferability, which is benefited from downscaling logits via L2-normalization.
Experiments conducted on the ImageNet dataset validate the effectiveness of the proposed methods, which outperform the state-of-the-art methods in black-box targeted attacks. The source code is available at \href{https://github.com/WJJLL/Target-Attack/}{Link}.
\end{abstract}

\begin{IEEEkeywords}
Adversarial machine learning, convolutional neural networks, Targeted adversarial examples, logit calibration
\end{IEEEkeywords}

\section{Introduction}
In the past decade, deep neural networks (DNNs) have achieved remarkable success in various fields, \textit{e.g.}, image classification~\cite{simonyan2014very}, image segmentation~\cite{long2015fully}, and object detection~\cite{ren2015faster}. However, Goodfellow et al.~\cite{goodfellow2014explaining} revealed that the DNNs are vulnerable to adversarial attacks, in which adding imperceptible disturbances into the input can lead the DNNs to make an incorrect prediction. The adversarial perturbation will raise a vital threat for the real-world applications of the CNNs, especially in scenarios of speech recognition~\cite{wang2020towards,wu2020audio}, facial verification systems~\cite{zhong2020towards,dong2019efficient} and person re-identification systems~\cite{ding2021beyond,yang2022towards}. Many following 
approaches~\cite{dong2018boosting,dong2019evading,cohen2019certified,tramer2017ensemble,xie2019improving,byun2022improving,li2022decision} have been proposed to construct more destructive adversarial samples for investigating the vulnerability of the DNNs. Besides, some studies~\cite{goodfellow2014explaining,liu2016delving} also showed that the adversarial samples are transferable across different networks, raising a more critical robustness threat under the black-box scenarios. Therefore, it is vital to explore the vulnerability of the DNNs, which is extremely useful for designing robust DNNs.

Currently, most of the works~\cite{dong2018boosting,xie2019improving,lin2019nesterov,huang2019enhancing,wu2020skip,guo2020backpropagating,xu2022bounded} have been devoted to the untargeted black-box attacks, in which adversarial examples are crafted to fool unknown CNN models predicting unspecified incorrect labels. For example, \cite{dong2018boosting,xie2019improving} leveraged input-level transformation or augmentation to improve the non-targeted transferability. \cite{huang2019enhancing} proposed a powerful intermediate feature-level attack. \cite{wu2020skip,guo2020backpropagating} demonstrated that backpropagating more gradients through the skip-connections can increase the transferability. \cite{xiong2022stochastic} proposed the stochastic variance reduced ensemble (SVRE) attack to reduce the gradient variance of the ensemble models for improving transferability.
Despite the success in non-targeted cases, targeted transferability remains challenging, which requires eliciting the black-box models into a pre-defined target category.

\begin{figure*}
  \centering
  \begin{subfigure}{0.32\linewidth}
     \includegraphics[width=\linewidth]{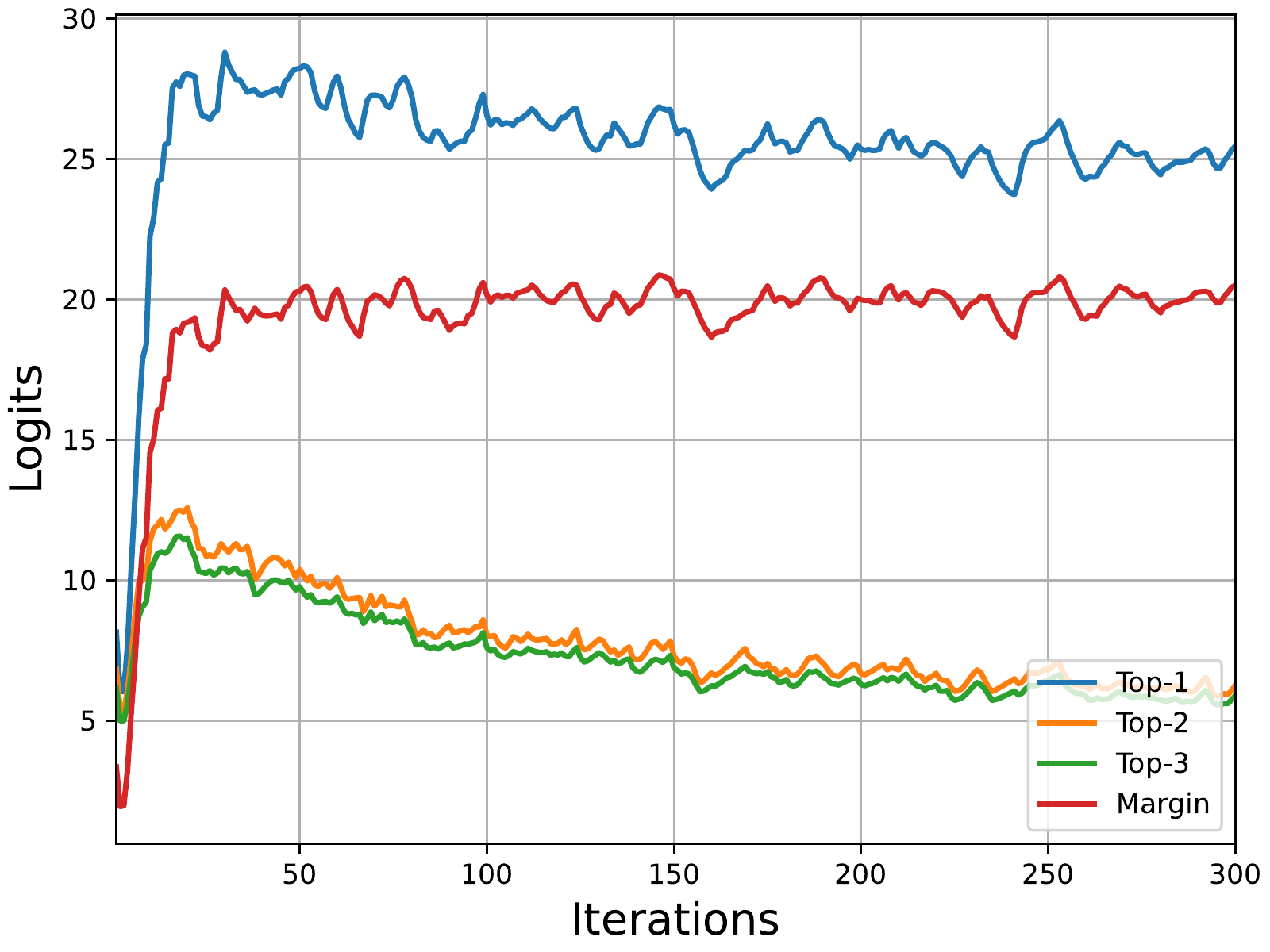}
    \caption{\label{fig:ce-1}Cross-Entropy}
  \end{subfigure}
  \begin{subfigure}{0.32\linewidth}
     \includegraphics[width=\linewidth]{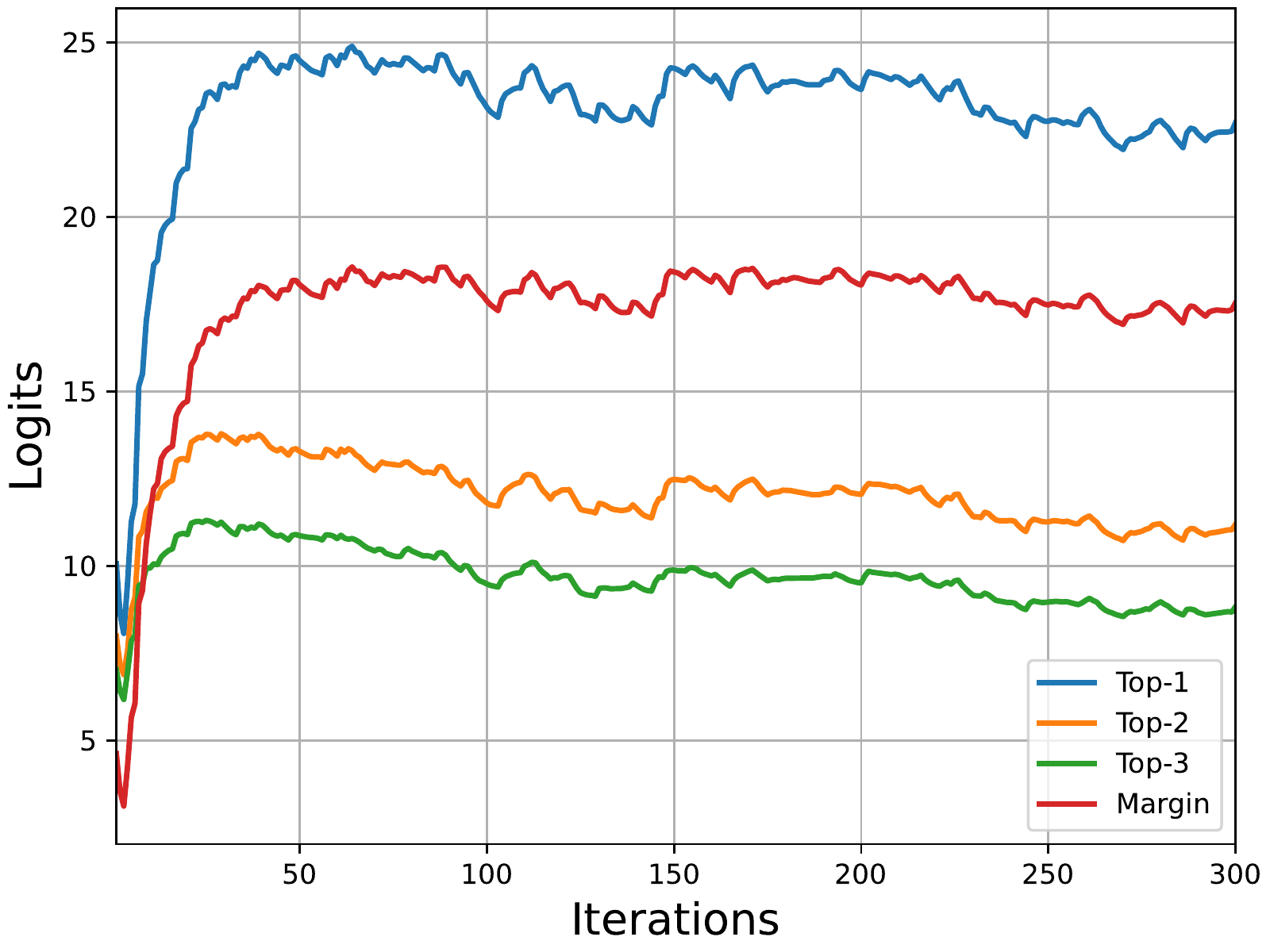}
    \caption{  \label{fig:po-trip-1}Po+Trip}
  \end{subfigure}
  \begin{subfigure}{0.32\linewidth}
     \includegraphics[width=\linewidth]{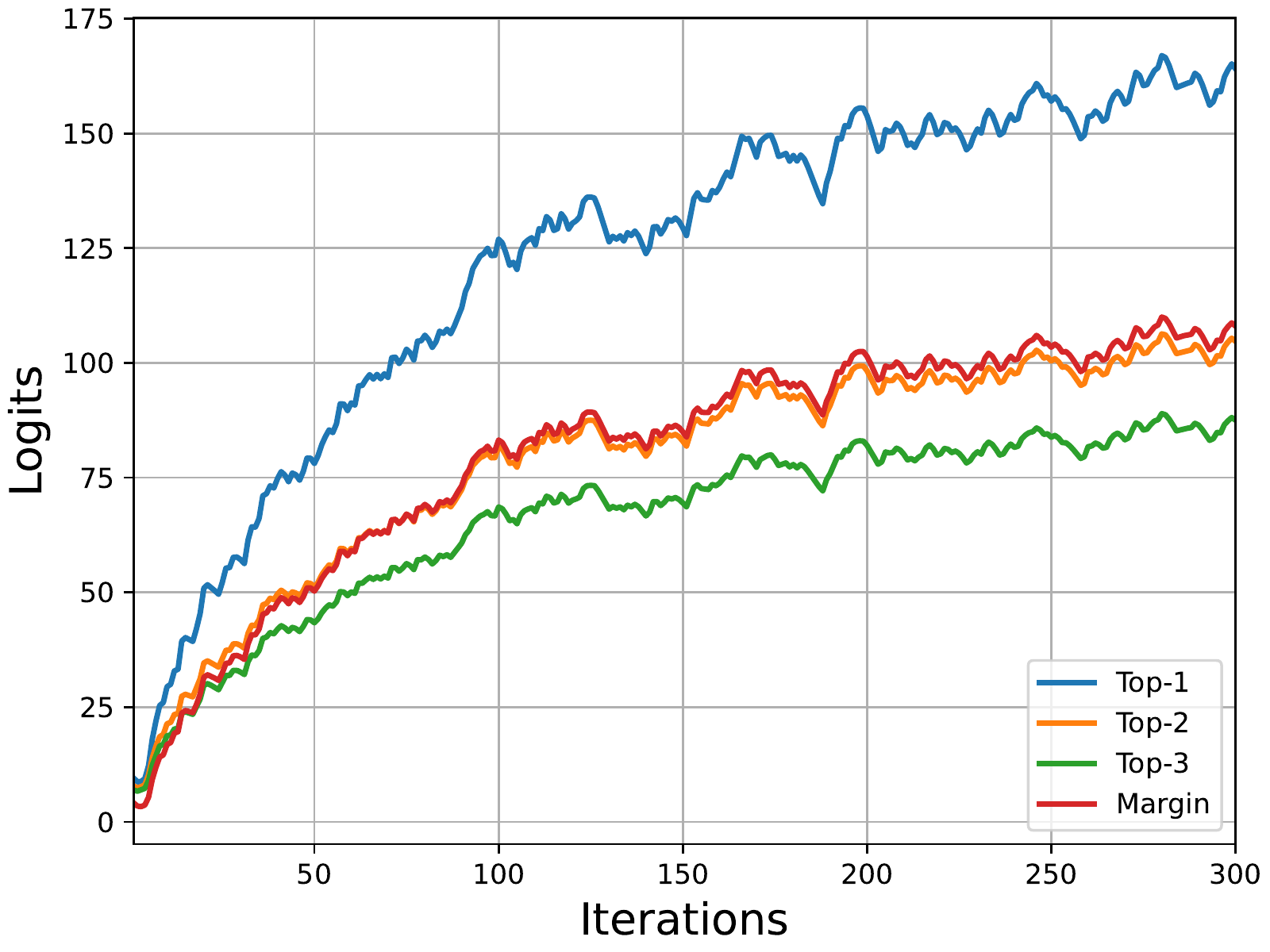}
    \caption{   \label{fig:logit-1} Logit}
  \end{subfigure}
  \caption{The average Top-3 logits and logit margin of 50 adversarial samples generated from the ResNet-50 
  by the Cross-Entropy, Po+Trip~\cite{li2020towards} and Logit~\cite{zhao2021success} loss functions. (*~Training and computation details of this figure are in Section~\ref{sec:logit_margin})}
    \label{fig:logits}
\end{figure*}

For learning the transferable adversarial samples in untargeted cases, most methods have leveraged the Cross-Entropy (CE) as the loss function. However, recent studies~\cite{li2020towards,zhao2021success} showed that the CE loss is insufficient for learning the adversarial perturbation in the targeted case due to the issue of vanishing gradient. To deal with this issue, Li et al.~\cite{li2020towards} adopted the Poincar{\'e} distance to increase the gradient magnitude adaptively during the optimization.
Zhao et al.~\cite{zhao2021success} demonstrated that an effortless logit loss equal to the negative value of the targeted logits could alleviate the gradient vanishing issue and achieve surprisingly strong targeted transferability.
Besides, Zhao et al.~\cite{zhao2021success} also showed that optimizing with more iterations can significantly increase the targeted transferability. 
Although it demonstrated that continually enlarging the logit of the targeted class along the whole training
iteration (as shown in Figure~\ref{fig:logit-1}) can improve the transferability of adversarial samples, it still does not thoroughly analyze the insufficient issue in the CE loss function

Different from ~\cite{li2020towards,zhao2021success}, in this study, we take a closer look at the vanishing gradient issue in the CE loss and find the logit margin between the targeted and non-targeted classes will quickly get saturated during the optimization {(as shown in Figure~\ref{fig:ce-1})}.
Moreover, this issue will influence the attack performance of the adversarial examples and thus essentially limits the transferability. Specifically, along with the training iterations in CE, we observe the logits of the targeted and non-targeted classes increase rapidly in the first few iterations. However, after reaching the peak, the logit margin between the targeted and non-targeted classes will get saturated, and further training will decrease the logits simultaneously to maintain this margin. This phenomenon is mainly due to the fact that the softmax function in CE will approximately output the probability of the target class to 1 when reaching the saturated margin (\textit{e.g.}, 10). Thus, it raises the problem that the transferability will not be further increased even with more optimization iterations.
A similar saturated phenomenon also can be observed in the Po+Trip loss~\cite{li2020towards} as shown in Figure~\ref{fig:po-trip-1}.
While in practice, we are encouraged to improve the transferability by keeping increasing the logit for the targeted class and its margin along the optimization to avoid saturation
against other non-targeted classes to cross the decision boundaries of other black-box models.  

In this paper, we devote to keeping enlarging logit margins along the optimization to alleviate the above saturation issue in CE. Inspired by the temperature-scaling used in knowledge distillation~\cite{hinton2015distilling}, a higher temperature $T$ will produce a softer probability distribution over different classes. We first utilize the scaling technique for targeted attacks to calibrate the logits. Then the logit margin between the targeted and non-targeted classes will not be saturated after only a few iterations and will keep improving the transferability. Besides, instead of using a constant $T$, we further explore an adaptive margin-based calibration by scaling the logits based on the logit margin of the target class and the highest non-target class. In addition, we also investigate the effectiveness of calibrating the targeted logit into the unit length feature space by L2-normalization, which is equivalent to minimizing the angle between the adversarial examples and the classifier weights of the targeted class. 

Finally, we conduct experiments on the ImageNet dataset to validate the effectiveness of the logits calibration for crafting transferable targeted adversarial examples. Experimental results demonstrate that the calibration of the logits helps achieve a higher attack success rate than other state-of-the-art methods. Besides, the combination of different calibrations can further provide mutual benefits.


\section{Related Works}
In this section, we give a brief introduction of the related works from the following two aspects: \textit{untargeted black-box attacks} and \textit{targeted attacks}.

\subsection{Untargetd Black-box Attacks}
After Szegedy et al.~\cite{szegedy2015going} exposed the vulnerability of deep neural networks, many attack methods~\cite{xie2019improving,dong2019evading} have been proposed to craft highly transferable adversaries in the non-targeted scenario. 
We first review several gradient-based attack methods that focus on enhancing the transferability against black-box models. 

\textbf{Iterative-Fast Gradient Sign Method (I-FGSM)}~\cite{kurakin2018adversarial} is an iterative version of FGSM~\cite{goodfellow2014explaining}, which iteratively adds the perturbation with a small step size $\alpha$ in the gradient direction:
\begin{equation}
   {\hat{x}}_{0} = x, \quad {\hat{x}}_{i+1} = {\hat{x}}'_{i} + \alpha \cdot \text{sign}(\nabla_{\hat{x}} J({\hat{x}}'_{i} , y)),
\end{equation}
where $\hat{x}_{i}'$ denotes the adversarial image in the $i_{th}$ iteration,  $\alpha=\epsilon/T$ ensures the adversary is constrained within an upper-bound perturbation $\epsilon$ through the $l_p$-norm when optimized by $T$ iterations. 

Following the seminal I-FGSM~\cite{kurakin2018adversarial}, a series of methods have been proposed to improve the transferability of attacking black-box models from different aspects, \textit{e.g.}, gradient-based, input augmentation-based. For example, the \textbf{Momentum Iterative-FGSM (MI-FGSM)}~\cite{dong2018boosting} introduces a momentum term to compute the gradient of the I-FGSM, encouraging the perturbation is updated in a stable direction. 
The \textbf{Translation Invariant-FGSM (TI-FGSM)}~\cite{dong2019evading} adopts a predefined kernel $W$ to convolve the gradient $\nabla_{\hat{x}} J({\hat{x}}'_{i}, y)$ at each iteration $t$, which can approximate the average gradient over multiple randomly translated images of the input $\hat{x}_t$. 
On the other aspects,
the \textbf{Diverse Input-FGSM (DI-FGSM)}~\cite{xie2019improving} leverages the random resizing and padding to augment the input $\hat{x}_t$ at each iteration. Currently, most targeted attack methods~\cite{li2020towards,zhao2021success,naseer2021generating} simultaneously use the MI~\cite{dong2018boosting}, TI~\cite{dong2019evading} and DI~\cite{xie2019improving} to form a strong baseline with better transferability.

\subsection{Targeted Attacks}
Targeted attacks are different from non-targeted attacks, which need to change the decision to a specific target class. ~\cite{kurakin2016adversarial} integrates the above non-targeted attack methods into targeted attacks to craft targeted adversarial examples. However, the performance is limited because it is insufficient to fool the black-box model only by maximizing the probability of the target class in the CE loss.

\textbf{Po+Trip}~\cite{li2020towards} finds the insufficiency of CE is mainly due to the vanishing gradient issue. Then, ~\cite{li2020towards} leverages the Poincar\'e space as the metric space and further utilizes Triplet loss to improve targeted transferability by forcing adversarial examples toward the target label and away from the ground-truth labels. To further address this gradient issue, \textbf{Logits}~\cite{zhao2021success} adopts a simple and straightforward idea by directly maximizing the target logit to pull the adversarial examples close to the target class, which can be expressed as:
\begin{equation}
\label{eq:logit}
    L_{Logit} = -z_t(\boldsymbol{x}'),
\end{equation}
 where $z_t(\cdot)$ is the output logits of the target class. Based on the prior \textbf{Logits}~\cite{zhao2021success}, the ODI~\cite{byun2022improving} proposes the object-based diverse input (ODI) method to diversify the input for further improving targeted transferability.

On the other hand, many studies employ resource-intensive approaches to achieve targeted attacks, which train target class-specific models (auxiliary classifiers or generative models) on additional large-scale data. For example, the FDA~\cite{inkawhich2020transferable,inkawhich2020perturbing} uses the intermediate feature distributions of CNNs to boost the targeted transferability by training class-specific auxiliary classifiers to model layer-wise feature distributions. The GAP~\cite{poursaeed2018generative} trains a generative model for crafting targeted adversarial examples. Subsequently,~\cite{naseer2019cross} adopts a relativistic training objective to train the generative model for improving attack performance and cross-domain transferability. Recently, the TTP~\cite{naseer2021generating} utilizes the global and local distribution matching for training target class-specific generators for obtaining high targeted transferability. However, the TTP requires actual samples from the target class and brings expensive training costs. 
Different from the above methods, we introduce three simple and effective logit calibrations into the CE loss, which can achieve competitive performance without additional data and training.










\section{Method}

\subsection{Problem Definition} 
Given a white-box model $\mathbb{F}_s$ and an input image $x$ not from the targeted class $t$, our primary goal is to learn an imperceptible perturbation $\delta$ that can fool the $\mathbb{F}_s$ to output the target $t$ for $\hat{x}=x+\delta$. Besides, the prediction of $\hat{x}$ will also be $t$ when feeding into other unknown black-box models. The $l_\infty$-norm is usually used to constrain the perturbation $\delta$ within an upper-bound $\epsilon$, denoted as $||\delta||_\infty  \leq \epsilon$. 

For the surrogate model $\mathbb{F}_s$, we denote the feature of the final classification layer of the input $x$ as $\phi(x)$. The logit $z_i$ of the category $i$ is computed by $z_i = W_i^T \phi(x) + b_i$, where $W_i$ and $b_i$ are the classifier weights and bias for category $i$. The corresponding probability $p_i$ after the softmax function is calculated by $p_i = \frac{e^{z_i}}{\sum e^{z_j}}$.



\begin{figure}[t]
\centering
\includegraphics[width=0.75\linewidth]{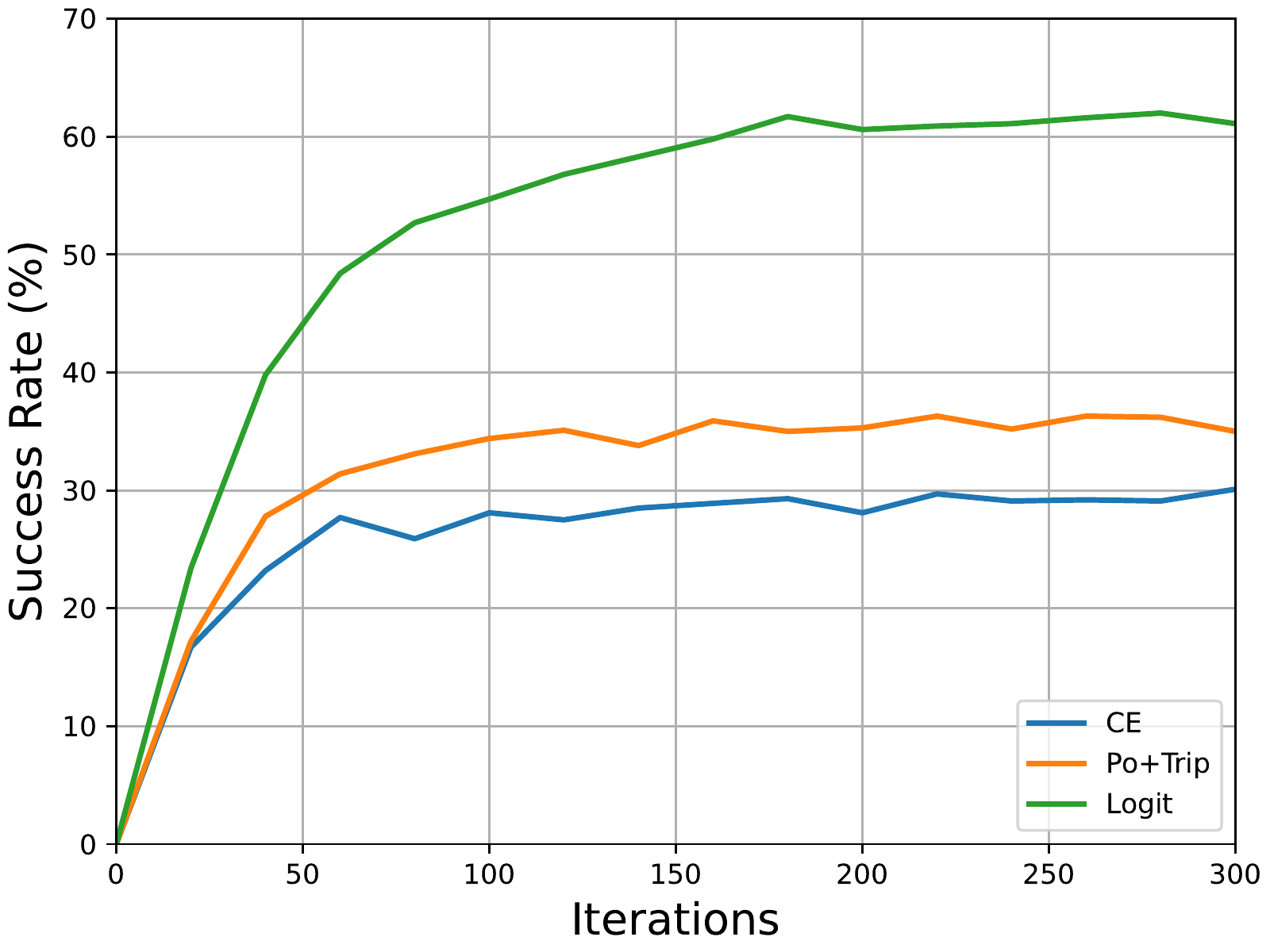}
\caption{The targeted attack success rate ($\%$) on VGG-16 by using the ResNet-50 as the surrogate model.}
\label{fig:res2vgg}
\end{figure}

\subsection{Logit Margin}
\label{sec:logit_margin}
When successfully attacking the $\mathbb{F}_s$, the logit $z_t$ of the target class $t$ will be higher than the logits $z_{nt}$ of any other non-target class in the classification task. Their logit margins can be computed by,
\begin{equation}
    G(\phi(\hat{x})) = z_t - z_{nt} = W_t^T \phi(\hat{x}) + b_t - W_{nt}^T \phi(\hat{x}) - b_{nt}.
    \label{eq:margin}
\end{equation}
\cite{li2020towards,zhao2021success} showed that it is insufficient to obtain transferable targeted adversarial samples that are only close to the target class while not far away enough from the true class and other non-targeted classes. Based on this property, it encourages us to continually enlarge this logit margin along the optimization to increase the separation between the targeted and other non-targeted classes, thereby improving transferability.

To have a better understanding of the relationship between the logit margins and the targeted transferability, we visualize the average Top-3 logits (1 targeted class and other two non-targeted classes) of 50 random adversarial samples trained on ResNet50 by the CE, Po+Trip~\cite{li2020towards}, and the Logit~\cite{zhao2021success} loss functions. We also compute the average logit margin of the targeted class against the Top-20 non-targeted classes.
The logit and the average logit margin are shown in Figure~\ref{fig:logits}, and the transfer targeted attack success rate of these three loss functions from ResNet50 to VGG16 is plotted in Figure~\ref{fig:res2vgg}.

From Figure~\ref{fig:logits}, we can observe that the logits of the targeted class and the Top-2 non-targeted classes increase rapidly in the first few iterations for the CE and Po+Trip loss, as well as their logit margins. When reaching the peak, the margin is saturated, and the logits start to decrease simultaneously to maintain the saturated margin. By comparing the CE and Po+Trip, the Po+Trip needs slightly more iterations to reach the saturated status and thus shows a marginal better transferability than CE, as shown in Figure~\ref{fig:res2vgg}. In comparison, the Logit loss function will keep increasing the logits of the targeted category and the logit margin. Hence, the Logit loss function shows a much better transfer targeted-attack success rate than CE and Po+Trip. On the other hand, the Logit loss also significantly increases the logits from other non-targeted classes when training with more iterations. 

To further analyze why the logit margin will quickly reach saturation in the CE loss function and explore the effectiveness of increasing the margin during training. In the following sections, we will revisit the cross-entropy loss function and introduce the logit calibration to achieve this goal.

\subsection{Revisiting the Cross-Entropy Loss}

Firstly, our objective is to maximize the logit margin in Eq.~\ref{eq:margin}. After computing the gradient w.r.t. to $\phi(\hat{x})$, we can get
\begin{equation}
    \frac{\partial G}{\partial\phi(\hat{x})} = W_t - W_{nt}.
\end{equation}
This gradient indicates that the adversarial feature $\phi(\hat{x})$ needs to move towards the target class while apart from those non-target classes. Next, we compute the gradient w.r.t. to $\phi(\hat{x})$ in the Cross-Entropy loss function
\begin{equation}
    L_{ce} = - \log(p_t) = -z_t + \log(\sum e^{z_j}),
\end{equation}
where  the $p_t$ is probability about target class $t$ and the $z_t$ is the logit of the category $t$. And we can get the gradient 
\begin{align}
\label{eq:ce}
    \frac{\partial L_{ce}}{\partial \phi(\hat{x})} & = - \frac{\partial z_t}{\partial \phi(\hat{x})} + \frac{1}{\sum e^{z_j}}\cdot \frac{\partial \sum e^{z_j}}{\partial \phi(\hat{x})} \\ \nonumber
    &  = -\frac{\sum e^{z_i}}{\sum e^{z_j}} \cdot \frac{\partial z_t}{\partial \phi(\hat{x})} +  \frac{1}{\sum e^{z_j}} \sum e^{z_i}\frac{ \partial z_i}{\partial \phi(\hat{x})} \\ \nonumber
    & = \sum \frac{e^{z_i}}{\sum e^{z_j}} \cdot (\frac{\partial z_i}{\partial \phi(\hat{x})} - \frac{\partial z_t}{\partial \phi(\hat{x})}) 
    \\  
    &= \sum -p_i (W_t - W_i), \nonumber
\end{align}
where $W_t$ is the classifier weights and bias for category $t$. From Eq.~\ref{eq:ce}, we actually find that the CE loss is designed to adaptively optimize the $\phi(\hat{x})$ towards $W_t$ and away from other $W_i$. However, 
after optimization with several iterations, the $p_i$ of the non-targeted class will soon approach to $0$ and then the influence of $W_t - W_i$ significantly vanishes.

\begin{figure}[t]
    \centering
    \includegraphics[width=0.7\linewidth]{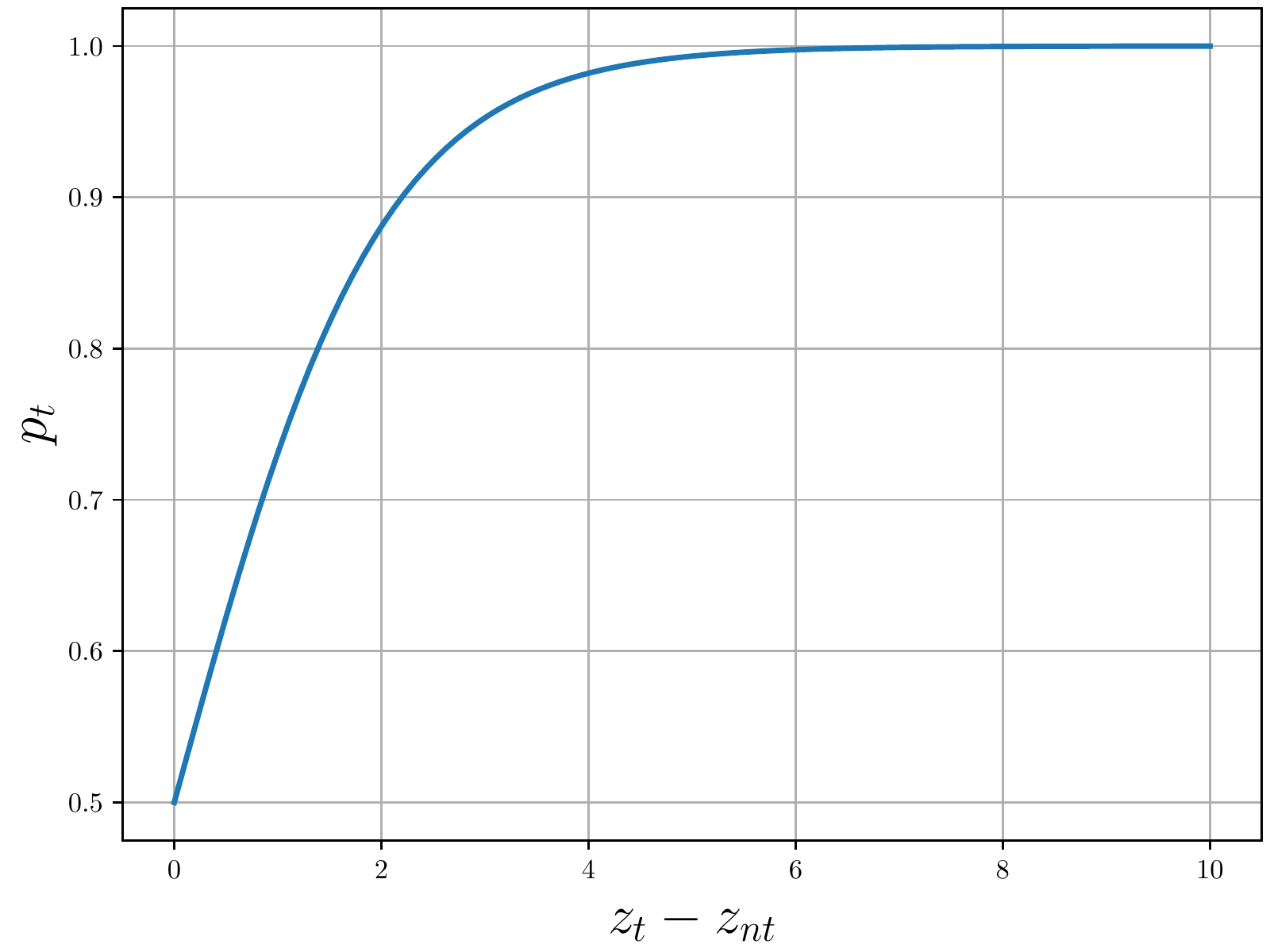}
    \caption{The probability of $p_t$ under different $z_t - z_{nt}$.}
    \label{fig:prob}
\end{figure}

Let's consider the case only with 2 classes ($t$ and $nt$), we have the probabilities $p_t$ and $p_{nt}$ as: 
\begin{equation}
    p_t = \frac{e^{z_t}}{e^{z_t} + e^{z_{nt}}} = \frac{1}{1 + e^{-(z_t-z_{nt})}},
\end{equation}
\begin{equation}
    p_{nt} = \frac{e^{z_{nt}}}{e^{z_t} + e^{z_{nt}}} = \frac{1}{1 + e^{(z_t-z_{nt})}}.
\end{equation}
As shown in Figure~\ref{fig:prob}, the $p_t$ will get close to 1 when $z_t - z_{nt}>6$ (\textit{e.g.}, $p_{nt}\approx 2e^{-9}$ when $z_t - z_{nt}=20$). 
In such a context, the gradient will significantly vanish. Recall that, in the CE loss function (Figure~\ref{fig:ce-1}), the logit margin increases rapidly, but will reach saturation when approaching a certain value. This further indicates that the optimization of the CE loss function is largely restrained when the logit margin reaches a certain value.
To this end, we raise the question \textit{if we explicitly enforce the optimization to increase the logit margin ($z_t - z_{nt}$) continually during the training process, could we get better transferable targeted adversarial samples?} 



To answer this, we propose to downscale the $z_t - z_{nt}$ by a factor $s$ in the CE and extend the informative optimization for more iterations. Since in such circumstance, $z_t - z_{nt}$ will be enlarged by the factor $s$. Specifically, suppose that the optimization will be saturated when $z_t - z_{nt}$ reaches a certain value $v$. Using $z_t - z_{nt}$ and $\frac{z_t - z_{nt}}{s}$ in the CE will both approach the saturated value $v$. Then, it is easy to infer that, for the latter case, $z_t - z_{nt}$ will be $v \times s$. 

\begin{figure*}[t]
    \centering
    \begin{subfigure}{0.32\textwidth}
        \includegraphics[width=\linewidth]{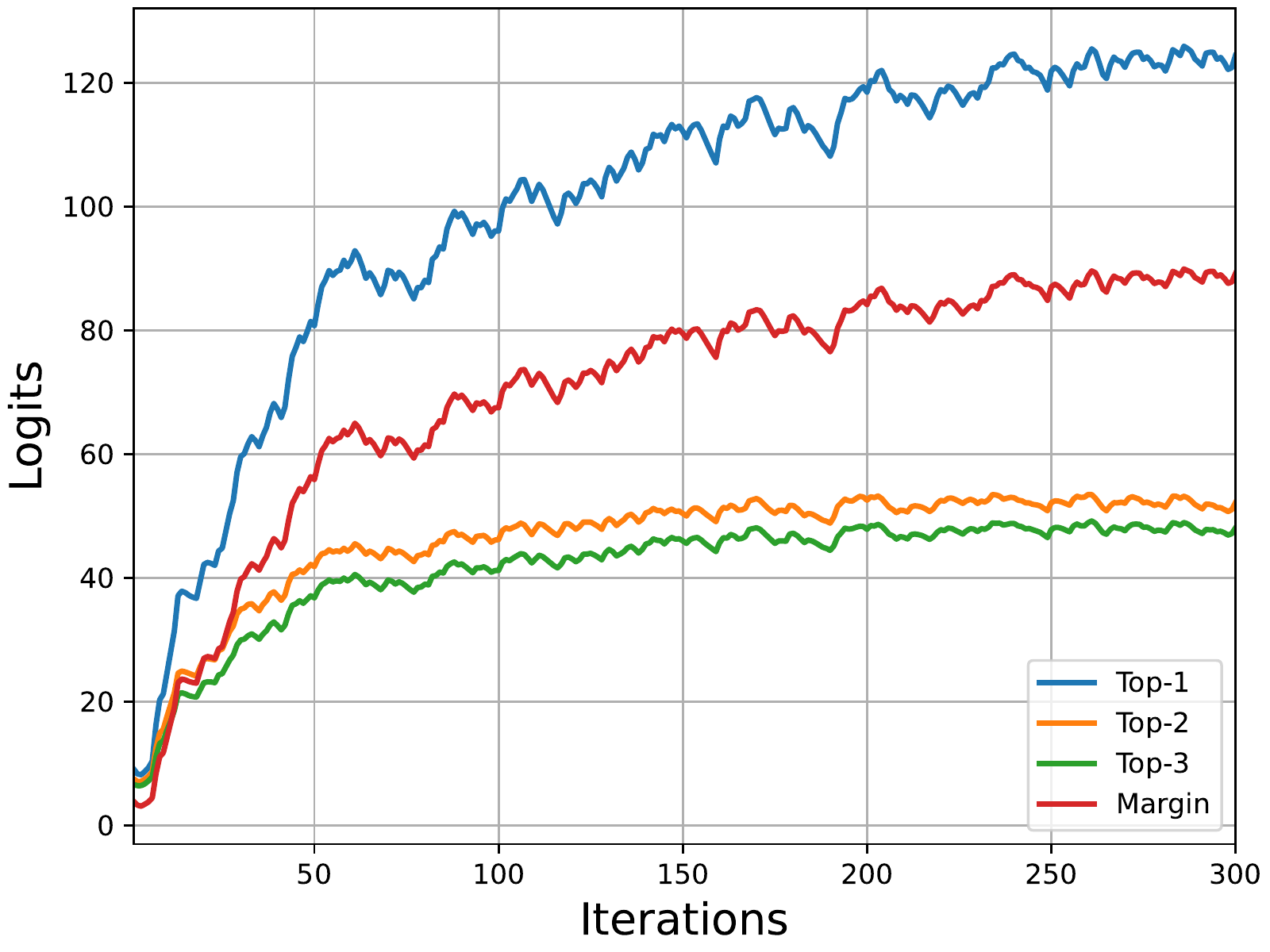}
        \caption{\label{fig:logits_cal-1}T=5}
    \end{subfigure}
    \begin{subfigure}{0.32\textwidth}
        \includegraphics[width=\linewidth]{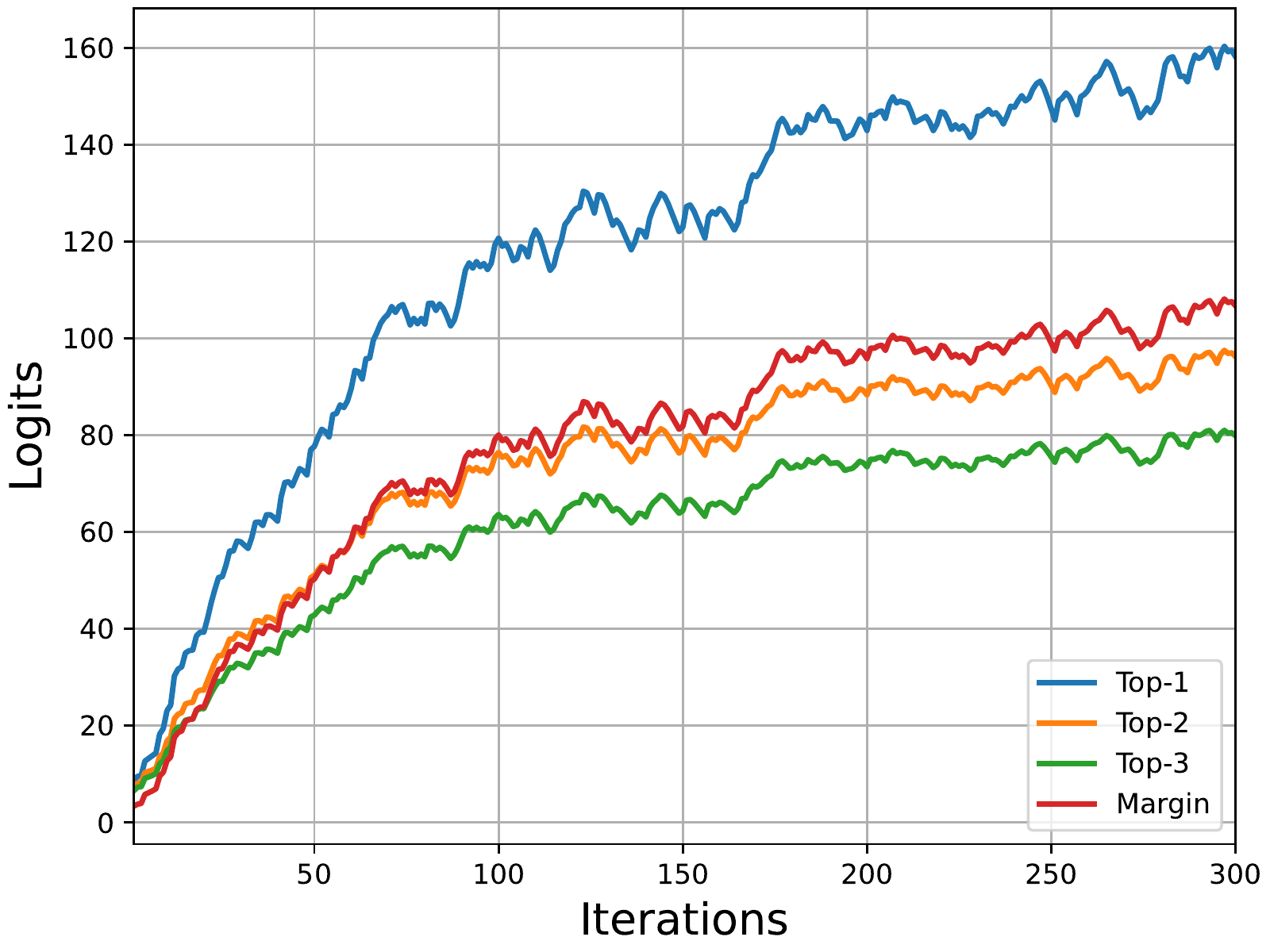}
        \caption{\label{fig:logits_cal-2}T=20}
    \end{subfigure}
    \begin{subfigure}{0.32\textwidth}
        \includegraphics[width=\linewidth]{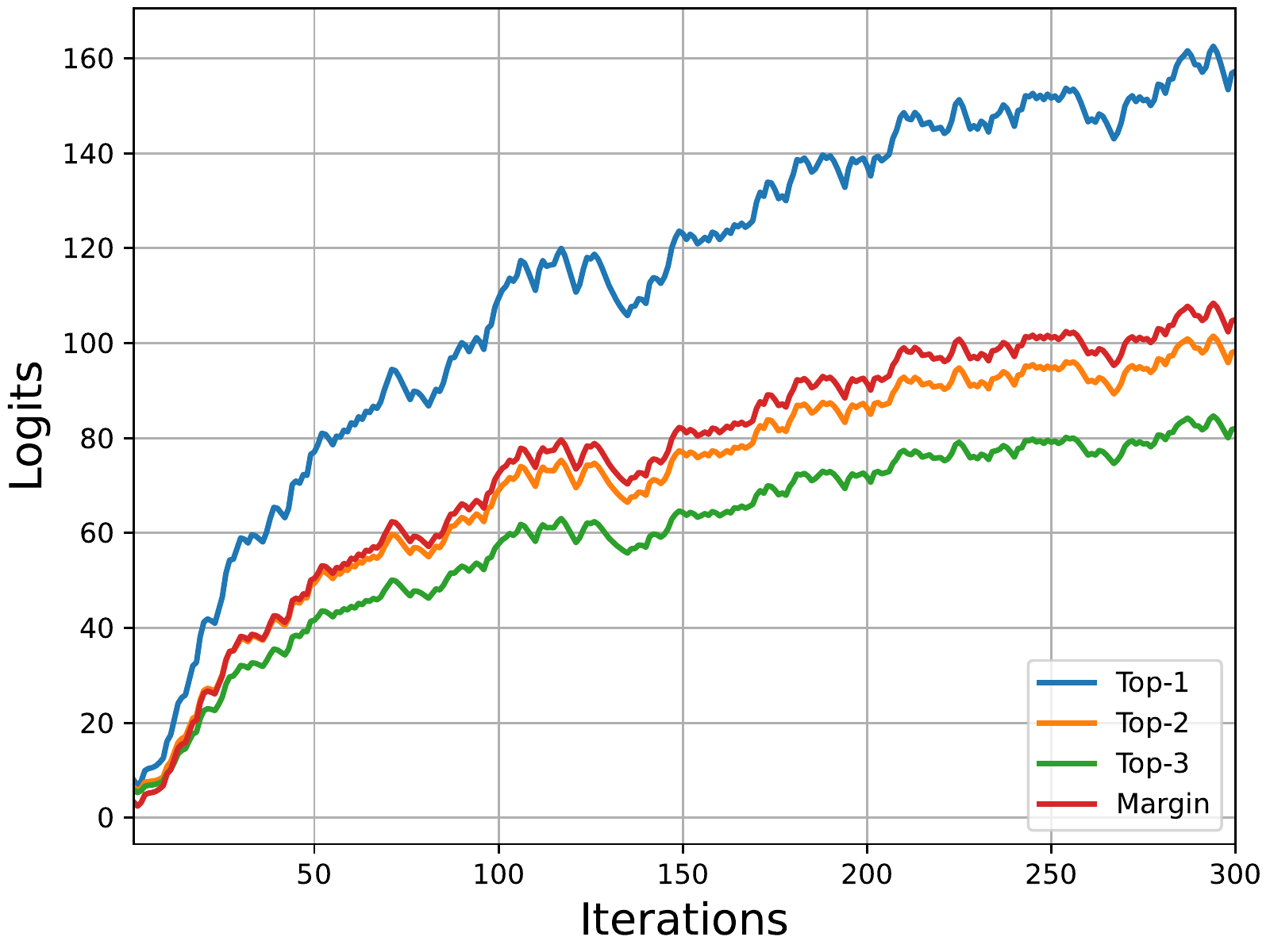}
        \caption{\label{fig:logits_cal-3}Margin}
    \end{subfigure}
    \caption{After the logit calibration, the average Top-3 logits and logit margin of 50 adversarial samples crafted from ResNet-50.}
    \label{fig:logits_cal}
\end{figure*}

\subsection{Calibrating the Logits}
To downscale the $z_t - z_{nt}$ during the optimization, we investigate three different types of logit calibration methods in this study, \textit{i.e.}, Temperature-based, Margin-based, and Angle-based.

\subsubsection{Temperature-based}

Inspired by the Temperature-scaling used in the knowledge distillation~\cite{hinton2015distilling}, our first logit calibration directly downscales the logits by a constant temperature factor $T$, 
\begin{equation}
    \Tilde{z}_i = \frac{z_i}{T}.
\end{equation}
After introducing the $T$, the probability distribution $\bm{p}$ will be softer over different classes. The corresponding gradient can be computed by:
\begin{align}
\label{eq:ce_deri}
     \frac{\partial L_{ce}^{T}}{\partial \phi(\hat{x})} &= \frac{e^{z_j / T}}{ \sum e^{z_j/T}} \cdot \frac{1}{T} (\frac{\partial z_j}{\partial \phi(\hat{x})} - \frac{\partial z_t}{\partial \phi(\hat{x})}) \\ \nonumber
     &= \sum -\hat{p_i}\frac{(W_t - W_i)}{T}.
\end{align}
After downscaling by the factor $T$, the new $\hat{p_i}$ after the softmax will not quickly approach 0 when only trained with a few iterations. In Figure~\ref{fig:logits_cal-1} and Figure~\ref{fig:logits_cal-2}, we visualize the trend of logits and margins using $T=5$ and $T=20$. We can find that targeted logits and the logit margin will keep increasing as the same as the Logit loss~\cite{zhao2021success} in Figure~\ref{fig:logits}. Meanwhile, the trend of $T=20$ is very similar to the Logit~\cite{zhao2021success}.




\subsubsection{Margin-based}
The previous Temperature-based logit calibration contains a hype-parameter $T$, which could be different for different surrogate models $\mathbb{F}_s$. To migrate this issue, we further introduce an adaptive margin-based logit calibration without extra hype-parameters. Specifically, we calibrate the logits by using the margin between the Top-2 logits in each iteration, denoted as:
\begin{equation}
    \Tilde{z}_i = \frac{z_i}{\hat{z}_{1} - \hat{z}_{2}},
\end{equation}
where $\hat{z}_{1}$ and $\hat{z}_{2}$ are the Top-1 and the Top-2 logit, respectively.
 Suppose the $\Tilde{\bm{z}}$ is sorted, the Top-1 logit $\Tilde{z}_1$ will be the target class  $\Tilde{z}_t$ after a few iterations. Therefore, the corresponding calibrated probability of the target class will be:
\begin{align}
    p_t &= \frac{1}{1 + \sum_{i\neq t} e^{-(\Tilde{z}_t-\Tilde{z}_i)}}  \\ \nonumber
        &= \frac{1}{1 + e^{-(\Tilde{z}_t-\Tilde{z}_i)} + \sum_{i=2} e^{-(\Tilde{z}_t-\Tilde{z}_i)}} \\ \nonumber
        &= \frac{1}{1 + e^{-\frac{\hat{z}_t-\hat{z}_2}{\hat{z}_t-\hat{z}_2}} + \sum_{i=2}e^{-(\Tilde{z}_t-\Tilde{z}_i)}}  \\ \nonumber
        &< \frac{1}{1 + e^{-1}}.
\end{align}

Correspondingly, we can have the probability of $1-p_{t} > 1- \frac{1}{1 + e^{-1}}$. Therefore, the probability $p_{\hat{1}}$ of the Top-1 non-target class will be larger than the average probability of all non-target classes, denoted as:
\begin{equation}
    p_{\hat{1}} =  \frac{1}{e^{\Tilde{z}_{\hat{1}}-\Tilde{z}_{t}} + \sum_{i\neq t} e^{\Tilde{z}_{i}-\Tilde{z}_{\hat{1}}}} > \frac{1}{N-1} (1-\frac{1}{1 + e^{-1}}).
\end{equation}
Then, it can adaptively deal with the vanishing gradient issue in the original CE loss function. The logits and the margin is shown in Figure~\ref{fig:logits_cal-3}. 

\begin{table*}[t]
\caption{The targeted transfer success rates (\%) in the single-model transfer scenario. (Results with 20/100/300 iterations are reported, and the highest one at 300 iterations is showed \textbf{bold}.)}
\centering
\begin{tabular}{l|ccc|ccc} \hline
\multirow{2}{*}{Attack} & \multicolumn{3}{c|}{Surrogate Model: ResNet50} & \multicolumn{3}{c}{Surrogate Model: Dense121} \\
                        &  $\rightarrow$Dense121       & $\rightarrow$VGG16    & $\rightarrow$Inc-v3    & $\rightarrow$Res50       & $\rightarrow$VGG16      & $\rightarrow$Inc-v3      \\ \hline
CE  & 27.0/40.2/42.7    &17.4/27.6/29.1    &2.3/4.1/4.6     &12.3/17.2/18.4     &8.6/10.5/10.9   &1.6/2.3/2.8     \\
Po+Trip  &27.9/51.2/54.8	&17.9/35.5/34.7	&3.2/6.8/7.8	&11.0/14.8/15.0	&7.3/9.2/8.6	&1.6/2.8/2.8      \\
Logit  &31.4/64.0/71.8	&23.8/55.0/62.4	&3.1/8.6/10.9	&17.4/38.6/43.5	&13.7/33.8/37.8	&2.3/6.6/7.5                                              \\ \hline
T=5   &33.3/69.9/\textbf{77.8} &24.8/59.9/66.1 &3.1/9.4/\textbf{12.2} &19.3/43.4/47.5 &14.6/36.6/39.4 &2.3/7.3/8.8 \\
T= 10 & 31.6/68.5/77.0 & 23.6/58.5/\textbf{66.4} & 2.8/9.4/11.6 & 17.9/43.2/\textbf{49.3} &	13.4/36.8/\textbf{41.5}	& 2.2/7.7/8.8\\
Margin & 33.3/65.8/76.5	&23.1/58.6/65.7	&3.0/9.5/12.2	&18.8/42.8/47.2	&14.5/36.5/41.4	& 2.5/7.7/\textbf{9.4} \\
Angle & 38.9/72.5/77.2 &	29.2/60.7/65.2	& 4.4/10.7/11.1 &	20.6/43.2/47.8	& 16.5/35.7/39.3	& 3.0/7.7/8.9  \\
\hline \hline
\multirow{2}{*}{Attack} & \multicolumn{3}{c|}{Surrogate Model: VGG16} & \multicolumn{3}{c}{Surrogate Model: Inc-v3} \\
                        & $\rightarrow$Res50 & $\rightarrow$Dense121           & $\rightarrow$Inc-v3    & $\rightarrow$Res50 & $\rightarrow$Dense121       & $\rightarrow$VGG16            \\ \hline
CE  & 0.5/0.3/0.6	&0.6/0.3/0.3	&0/0/0.1	&0.7/1.2/1.8	&0.6/1.3/1.9	&0.4/0.8/1.3\\
Po+Trip  & 0.7/0.6/0.7	& 0.7/0.6/0.5	& 0.1/0.1/0.1	& 1.0/1.6/1.7	& 0.6/1.7/2.5	& 0.7/1.2/1.8\\
Logit   &3.4/9.9/11.6	&3.5/12.0/13.9	&0.3/1.0/\textbf{1.3}	&0.6/1.1/2.0	&0.6/1.9/3.0	&0.6/1.5/2.8\\ \hline 
T=5   &3.1/7.0/6.9	&3.3/7.6/7.8	 &0.2/0.9/0.8	&0.7/1.7/2.1	&0.5/1.9/\textbf{3.3}	&0.4/1.6/2.6 \\
T= 10 &3.6/9.0/9.7	&3.4/10.5/11.7	&3.2/1.1/1.3	&0.5/1.3/1.9	&0.6/2.0/2.7	&0.4/1.5/\textbf{2.8}\\
Margin  & 3.3/10.3/\textbf{12.0}	&3.5/12.5/\textbf{14.5}	&0.3/1.1/\textbf{1.3}	&0.5/1.4/1.7	&0.7/2.1/3.1	& 0.5/1.7/2.7 \\
Angle & 0.4/0.7/0.5	& 0.6/0.4/0.5	&0/0/0.1	&0.8/1.8/\textbf{2.6}	&0.8/2.2/3.0	&0.9/1.7/2.4  \\\hline
\end{tabular}
\label{tab:soat}
\end{table*}

\subsubsection{Angle-based}
On the other aspect, we also can adaptively calibrate the logit $W_i^T \phi(\hat{x})+b_i$ of each category based on the weights $W_i$. Besides, the weights $W_i$ for different category $i$ usually has a different norm. To further alleviate the influence of various norms, we calibrate the logit into the feature space with unit length by L2-normalization, denoted as $\frac{W_i^T \phi(\hat{x})+b_i}{ ||W_i|| ||\phi(\hat{x})||}$. 

If omit the $b_i$, this calibration is actually computed the $\cos(\theta)$ between $\phi(\hat{x})$ and $W_i$, and we term it as angle-based calibration. \textit{Since this angle-based calibration will bound each logit smaller than one.} 
Instead of using the CE loss function, we directly minimize the angle between the $\phi(\hat{x})$ and the targeted weights $W_t$. The corresponding optimization loss function is:
\begin{equation}
    L_{cosine} = - \frac{W_t^T \phi(\hat{x})}{ ||W_t|| ||\phi(\hat{x})||}.
\end{equation}
Notice that, the angle-based classifiers have been widely used in Face-Recognition task~\cite{liu2017sphereface,deng2019arcface}. 

Additionally, the three different calibration methods used in this study are complementary to other, and we will evaluate their performance and their mutual benefits in the following experimental section.

\section{The connection with Logit Loss}

In this section, we theoretically analyze the relationship between the Logit loss~\cite{zhao2021success} and Temperature-based calibration with a large $T$.
When using a large $T$, the distribution ${\hat{p}_i}$ will be extremely smooth over different classes. And we can get the $\hat{p}_i \approx \frac{1}{N}$ for each class, where $N$ is the number of classes. Besides, the $\sum_{i}{W_i}$ are with very small values (i.e., $\sum_{i}{W_i}\approx 0$). In this study, we conduct experiments on the ImageNet dataset ($N=1000$). Then the gradient of Cross-Entropy with a large T in Eq.~\ref{eq:ce_deri} will become:
\begin{align}
    \label{eq:ce_t}
    \frac{\partial L_{ce}^{T}}{\partial \phi(\hat{x})} & \approx \sum_{i}{- \frac{(W_t - W_i)}{NT}}\\ \nonumber
    & \approx -\frac{W_t}{T} + {\frac{1}{NT}} \sum_{i}{W_i} \\ \nonumber
    & \approx -\frac{W_t}{T}.
\end{align}

For the Logit loss~\cite{zhao2021success}, we can obtain the gradient w.r.t. to input feature $\phi(\hat{x})$ in Eq.~\ref{eq:logit} as:
\begin{equation}
    \frac{\partial L_{Logit}}{\partial \phi(\hat{x})} = - W_t.
    \label{eq:logit_w}
\end{equation} 

From the Eq.~\ref{eq:ce_t} and Eq.~\ref{eq:logit_w}, we can observe the gradient of  Cross-Entropy with a large T is approximate $\frac{1}{T}$ of the gradient of the Logit loss~\cite{zhao2021success}. On the other aspect, the I-FGSM is used for optimization,
\begin{equation}
    {\hat{x}}_{i+1} = {\hat{x}}'_{i} + \alpha \cdot \text{sign}(\nabla_{\hat{x}} L({\hat{x}}'_{i}, y)),
\end{equation}
which only considers the sign of the gradient. Therefore, Eq.~\ref{eq:logit_w} and Eq.~\ref{eq:ce_t} will update the perturbation in a similar direction. Based on the above analysis, we can consider the Logit
loss function as a special case of Temperature-based calibration with a large $T$. Besides, we compare the performance between the Logit and CE (T=50 \& T=100) in the following experimental section.

\section{Experiments}
\label{sec:exp_set}
\subsection{Experimental Setup.} 

In this section, we evaluate the effectiveness of logit calibration for improving transferable targeted adversarial attacks. Following the recent study~\cite{zhao2021success}, we conduct the experiments on the difficult ImageNet-Compatible Dataset\footnote{\url{https://github.com/cleverhans-lab/cleverhans/tree/master/cleverhans_v3.1.0/examples/nips17_adversarial_competition/dataset}}. This dataset contains 1,000 images with 1,000 unique class labels corresponding to the ImageNet dataset~\cite{deng2009imagenet}. We implement our methods based on the source code\footnote{\url{https://github.com/ZhengyuZhao/Targeted-Tansfer}} provided by the Logit~\cite{zhao2021success}. The same four diverse CNN models are used for evaluation, \textit{i.e.}, ResNet-50~\cite{he2016deep}, DenseNet-121~\cite{huang2017densely}, VGG-16 with Batch Normalization~\cite{simonyan2014very} and Inception-v3~\cite{szegedy2016rethinking}. The perturbation is bounded by $L_\infty \leq 16$. The TI~\cite{dong2019evading}, MI~\cite{dong2018boosting} and DI~\cite{xie2019improving} are used for all attacks, and $||W||_1 =5$ is set for TI. The I-FSGM is adopted for optimization with the $\alpha=2$. The attacks are trained for 300 iterations on an NVIDIA-2080 Ti GPU. We run all the experiments 5 times and report the average targeted transfer success rates (\%). Besides, we also conduct more experimental results about the ensemble attack and the real-world attack.

\subsection{Comparison with Others in Single-Model Transfer}

We first compare the proposed (temperature-based, margin-based and angle-based) logit calibrations with the original CE, Po+Trip~\cite{li2020towards}, and Logit~\cite{zhao2021success} in the single-model transfer task. In this task, we take one surrogate model for training and test the targeted transferability in attacking the other 3 models.
As shown in Table~\ref{tab:soat}, the original CE loss function produces the worst performance than the Po+Trip and Logit. But after performing the logit calibration in the CE loss function, we can find a significant performance is boosted compared with the original CE. Our calibration methods can outperform the Logit, especially when using the ResNet50 and Dense121 as the surrogate model. These results indicate that continuously increasing the logit margin can significantly influence the performance of the targeted transferability. On the other aspect, we find that $T=10$ has better performance than $T=5$ on the VGG-16, suggesting that different models may need different $T$. Instead of finding the best $T$ for a different model, the Margin-based calibration can solve the issue and reach the overall best transferability in all four models. However, we find that the Angle-based calibration is not working on the VGG16, which needs further investigation.

\begin{table*}[t]
\caption{The targeted transfer success rates (\%) by using different $T$ in CE loss function. (Results with 20/100/300 iterations are reported.)}
\centering
\begin{tabular}{l|ccc|ccc} \hline
\multirow{2}{*}{Attack} & \multicolumn{3}{c|}{Surrogate Model: ResNet50} & \multicolumn{3}{c}{Surrogate Model: Dense121} \\
                        &  $\rightarrow$Dense121       & $\rightarrow$VGG16    & $\rightarrow$Inc-v3    & $\rightarrow$Res50       & $\rightarrow$VGG16      & $\rightarrow$Inc-v3      \\ \hline

Logit  &31.4/64.0/71.8	&23.8/55.0/62.4	&3.1/8.6/10.9	&17.4/38.6/43.5	&13.7/33.8/37.8	&2.3/6.6/7.5 \\                    
T=0.5 &  13.2/16.0/19.5 & 7.1/9.5/11.0 & 1.2/1.8/2.4 &4.2/5.0/6.2 &2.5/3.5/3.2 &0.6/0.9/1.1\\
T=1   & 27.0/40.2/42.7    &17.4/27.6/29.1    &2.3/4.1/4.6     &12.3/17.2/18.4     &8.6/10.5/10.9   &1.6/2.3/2.8 \\
T=2 &34.2/62.8/67.7	&24.4/52.3/53.9	&3.3/7.2/8.5	&18.7/35.0/36.1	&13.2/27.3/27.0	&2.2/5.5/6.1\\
T=5   &33.3/69.9/\textbf{77.8} &24.8/59.9/66.1 &3.1/9.4/\textbf{12.2} &19.3/43.4/47.5 &14.6/36.6/39.4 &2.3/7.3/8.8  \\
T=10 & 31.6/68.5/77.0 & 23.6/58.5/\textbf{66.4} & 2.8/9.4/11.6 & 17.9/43.2/\textbf{49.3} & 13.4/36.8/\textbf{41.5}	& 2.2/7.7/\textbf{8.8}\\
T=20 & 30.4/65.6/74.3	& 22.9/55.4/63.6	& 3.2/9.0/11.6	& 17.6/40.3/46.2	& 13.4/35.4/40.1	& 2.3/6.7/8.7\\

T=50   &30.2/64.7/72.7	&23.3/55.1/62.9	&2.9/8.8/11.4	&17.3/39.6/44.8	&12.7/34.3/38.3	&2.4/6.7/8.3 \\
T=100 & 30.0/64.7/72.3	&22.8/54.4/61.9	&3.1/8.7/10.7	&17.0/39.7/44.7	&13.0/33.7/39.1	&2.2/6.5/8.1\\

\hline \hline
\multirow{2}{*}{Attack} & \multicolumn{3}{c|}{Surrogate Model: VGG16} & \multicolumn{3}{c}{Surrogate Model: Inc-v3} \\
                        & $\rightarrow$Res50 & $\rightarrow$Dense121           & $\rightarrow$Inc-v3    & $\rightarrow$Res50 & $\rightarrow$Dense121       & $\rightarrow$VGG16            \\ \hline

Logit   &3.4/9.9/11.6	&3.5/12.0/13.9	&0.3/1.0/1.3	&0.6/1.1/2.0	&0.6/1.9/3.0	&0.6/1.5/2.8\\ 
T=0.5   &0.2/0.1/0.2	&0.1/0.1/0.1	&0/0/0	&0.3/0.9/0.9	&0.3/0.8/1.4	&0.3/0.6/1.3\\
T=1   &0.5/0.3/0.6	&0.6/0.3/0.3	&0/0/0.1	&0.7/1.2/1.8	&0.6/1.3/1.9	&0.4/0.8/1.3 \\
T=2  &1.6/1.8/1.8	&1.8/1.9/1.6	&0.2/0.2/0.2	&0.6/1.5/2.0	&0.4/1.7/2.2	&0.5/1.2/2.0\\
T=5   &3.1/7.0/6.9	&3.3/7.6/7.8	&0.2/0.9/0.8	&0.7/1.7/2.1	&0.5/1.9/3.3	&0.4/1.6/2.6  \\
T=10 &3.6/9.0/9.7	&3.4/10.5/11.7	&0.3/1.1/1.3	&0.5/1.3/1.9	&0.6/2.0/2.7	&0.4/1.5/\textbf{2.8}\\
T=20 &3.4/9.7/\textbf{11.1}	&3.6/12.7/\textbf{13.8}	&0.3/1.2/\textbf{1.3}	&0.5/1.4/\textbf{2.3}	&0.6/1.8/\textbf{3.1}	&0.5/1.6/2.4\\

T=50   &3.1/10.2/11.4	&3.9/12.0/14.5	&0.1/1.1/1.3	&0.6/1.8/2.1	&0.6/2.0/3.0	&0.3/1.7/2.7 \\
T=100 &3.6/9.8/11.3	&3.4/11.8/13.9	&0.4/1.2/1.4	&0.6/1.6/2.0	&0.4/2.1/3.0	&0.4/1.7/2.8\\

\hline
\end{tabular}
\label{tab:ce_t}
\end{table*}

\subsection{Influence of Different $T$ in CE}
\begin{figure*}
  \centering
  \begin{subfigure}{0.32\textwidth}
     \includegraphics[width=\linewidth]{Figure/Logits_logit.pdf}
    \caption{Logit}
  \end{subfigure}
  \begin{subfigure}{0.32\textwidth}
     \includegraphics[width=\linewidth]{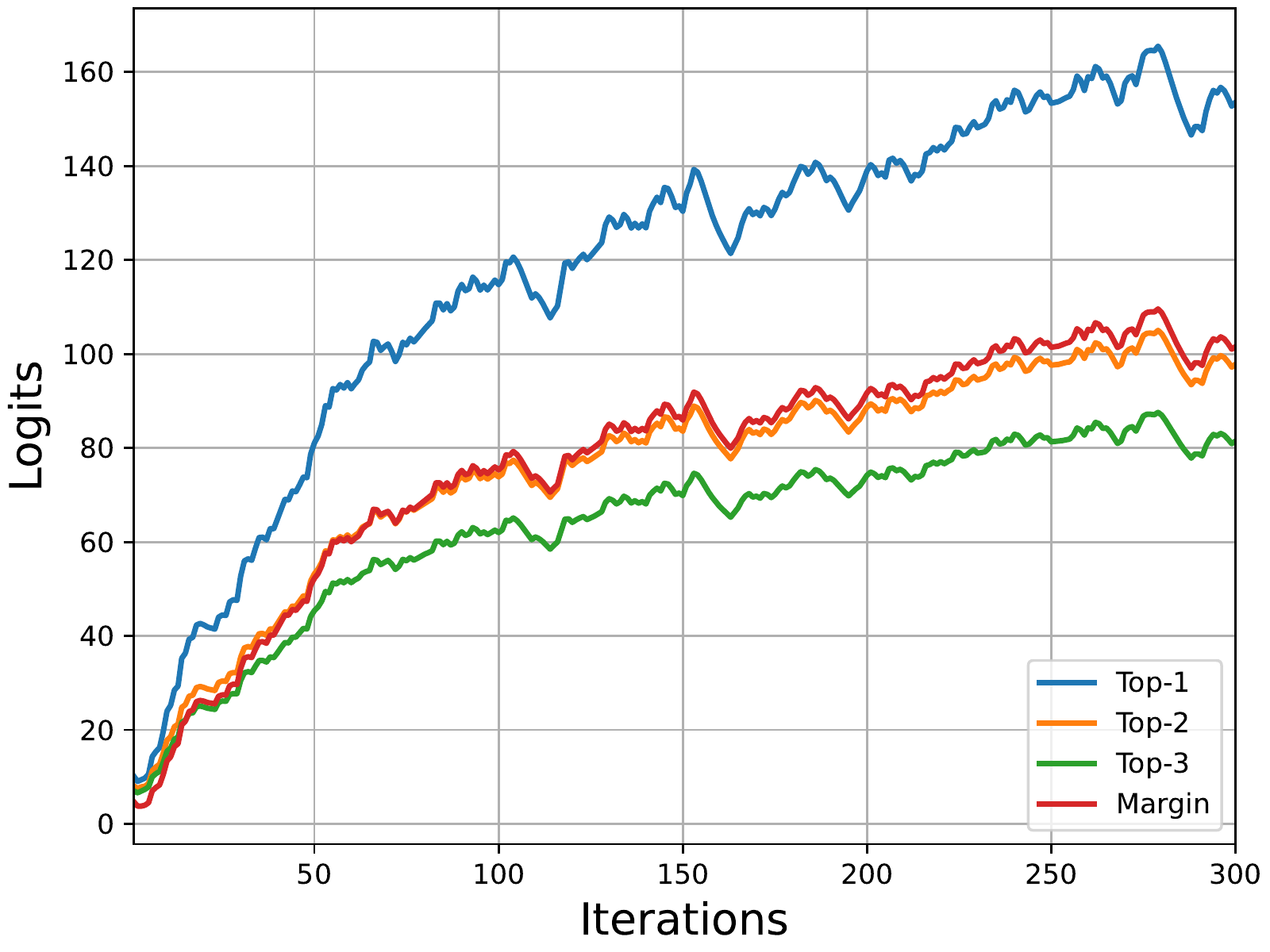}
    \caption{CE (T=50)}
  \end{subfigure}
  \begin{subfigure}{0.32\textwidth}
     \includegraphics[width=\linewidth]{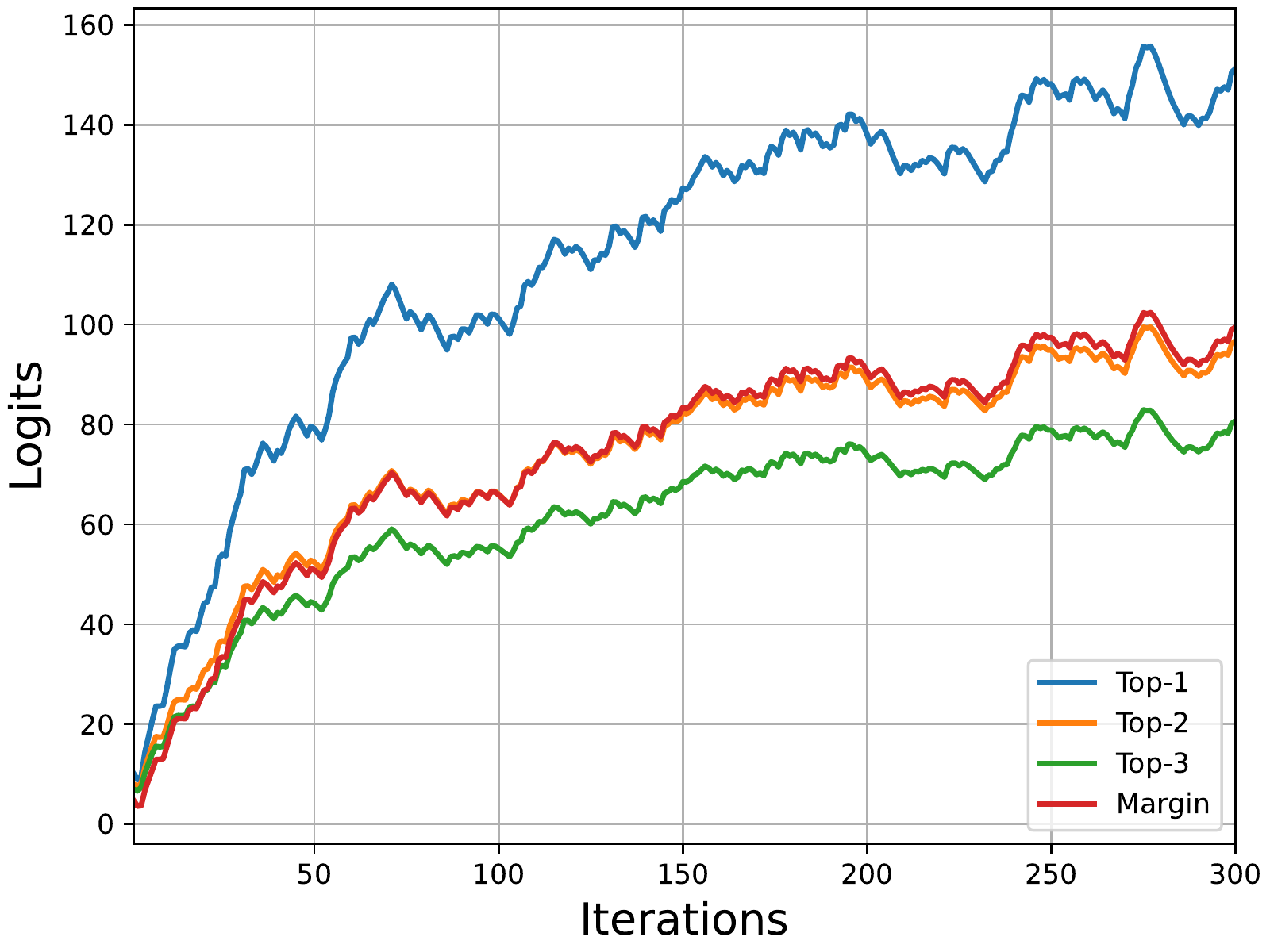}
    \caption{CE (T=100)}
  \end{subfigure}
    \caption{The average Top-3 logits and logit margin of 50 adversarial samples trained by the Logit, CE (T=50) and CE (T=100) loss functions for crafting the ResNet-50.}
    \label{fig:logit_ce_t}
\end{figure*}

In this section, we evaluate the influence of using different T in the CE loss function. The results are reported in Table~\ref{tab:ce_t}. From the Table, we can have the following observations. 
\textbf{1)} The scaling factor $T$ has a significant influence on the targeted transferability. Specifically, the performance drops significantly when using a small $T=0.5$. After increasing the $T$, we can observe the number of successfully attacked samples will increase.
\textbf{2)} The optimal $T$ for different model is different. For example, $T=5$ can produce the overall best performance for ResNet50, Dense121, and Inception v3, {while the VGG16 with fewer convolutional layers requires a large $T$ to obtain better transferability.} 
\textbf{3)} The performances are comparable when using $T=5$ and $T=10$ for ResNet50, Dense121, and Inception v3. This is because we use I-FSGM for optimization, which only considers the sign of the gradients.
\textbf{4)} When using a large $T$, the performance will be
similar to the Logit loss. 
For example, we compare the Logit with CE ($T=50$ $\&$  $T=100$) in the single-model transfer scenario, and we can see the performances are approximate for the targeted transfer success. Besides, in Figure~\ref{fig:logit_ce_t}, we visualize the trend of logits and margins using the Logit loss~\cite{zhao2021success} and the CE loss (T=50 $\&$ T=100). We can find the trends are very similar when the adversarial samples 
are crafted from the ResNet-50 by the Logit and CE ($T=50$ $\&$  $T=100$) loss functions.

\subsection{Mutual Benefits of Different Logit Calibrations}

In this part, we evaluate the mutual benefits of combining different calibrations. As shown in Table~\ref{tab:mutual}, we can have the following findings. \textbf{1)} \textcolor{black}{Combining the T=5/10/20 and Margin, there is no increase in performance compared with using one of them.} This is because the gradient directions of these two methods are very similar. \textbf{2)} Combining the T=5 and Angle, we can observe a further improvement when using ResNet50 and Dense121 as the surrogate model, \textit{e.g.}, the transferable rate of ``ResNet50 $\rightarrow$ Dense121'' is increased to 82.4\% with 300 iterations. \textcolor{black}{Since the Angle obtains poor performance on VGG16, the transferable rates of corresponding combinations are also low in T=5/10+Angle, but T=20+Angle can deal with this issue.} \textbf{3)} Combining the Margin and Angle, there are only slight improvements on ResNet50 and Dense121, while it can alleviate the negative effects caused by the angle-based calibration.

Finally, by jointly considering the results in Table~\ref{tab:soat}, ~\ref{tab:ce_t} and ~\ref{tab:mutual}, we suggest that the single Margin-based calibration is effective for improving the transferability for all different CNNs. Besides, it is also preferred to use T=5 + Angle for further improving the performance for CNN with more layers, \textit{e.g.}, ResNet50 and Dense121.

\begin{table*}[t]
\caption{The comparison of combining logit calibrations. (The targeted transfer success rates (\%) with 20/100/300 iterations are reported.)}
\label{tab:mutual}
\centering
\begin{tabular}{l|ccc|ccc} \hline
\multirow{2}{*}{Attack} & \multicolumn{3}{c|}{Surrogate Model: ResNet50} & \multicolumn{3}{c}{Surrogate Model: Dense121} \\
                        &  $\rightarrow$Dense121       & $\rightarrow$VGG16    & $\rightarrow$Inc-v3    & $\rightarrow$Res50       & $\rightarrow$VGG16      & $\rightarrow$Inc-v3      \\ \hline
T=5 + Margin  &33.8/69.8/77.2	&24.0/59.0/65.5	&3.3/9.6/11.1	&19.3/44.3/47.8	&14.1/37.7/40.8	&2.5/7.5/9.4   \\
T=5 + Angle   &34.5/74.3/\textbf{82.4}	& 25.6/66.5/\textbf{72.2}	& 3.6/10.5/\textbf{13.1}	& 20.3/52.7/\textbf{61.9}	& 15.8/45.0/\textbf{53.6}	& 2.3/9.2/\textbf{12.7} \\
T=10 + Margin	&32.7/69.5/77.3	&22.8/59.4/66.3	&12.9/9.7/11.5 &18.3/44.1/49.1	&13.7/36.9/41.6	&2.4/8.3/9.2\\
T=10 + Angle	&33.0/69.8/79.1	&24.4/59.0/68.9	&3.4/10.0/12.9 &19.4/47.2/56.1	&14.8/40.1/47.0	&2.5/8.3/11.0\\
T=20 + Margin	&33.0/69.2/76.2	&23.1/58.4/65.8	&3.2/9.5/11.8  &19.1/43.4/48.5	&13.9/36.7/41.4	&2.4/7.8/9.5\\
T=20 + Angle	&34.2/68.6/76.5	&24.7/58.7/66.6	&3.4/9.7/12.7  &20.0/44.4/50.9	&15.5/38.4/43.7	&2.5/8.2/9.5\\
Margin + Angle  &34.4/70.8/78.1	&24.3/60.2/67.4	&3.5/10.4/12.6 &19.9/46.6/52.7	&15.2/39.3/44.5	& 2.7/8.2/9.9\\

\hline \hline
\multirow{2}{*}{Attack} & \multicolumn{3}{c|}{Surrogate Model: VGG16} & \multicolumn{3}{c}{Surrogate Model: Inc-v3} \\
                        & $\rightarrow$Res50 & $\rightarrow$Dense121           & $\rightarrow$Inc-v3    & $\rightarrow$Res50 & $\rightarrow$Dense121       & $\rightarrow$VGG16            \\ \hline
T=5 + Margin    &3.5/10.2/11.4	&3.7/12.4/14.6	&0.3/1.1/1.3	&0.5/1.4/1.6	&0.6/2.1/2.9	&0.5/1.7/2.8 \\
T=5 + Angle     &2.2/2.5/2.3	&2.4/2.6/2.3	&0.2/0.1/0.2 &0.5/1.6/\textbf{2.4}	&0.6/2.0/3.1	&0.5/1.7/2.5\\
T=10 + Margin	&3.2/10.7/11.7	&3.4/12.9/\textbf{15.0}	&0.2/1.0/\textbf{1.4} &0.5/1.4/1.9	&0.5/1.9/3.0	&0.3/1.5/2.3\\
T=10 + Angle	&3.4/6.2/5.1	&3.5/7.5/7.0	&0.2/0.6/0.6 &0.6/1.3/1.9	&0.6/2.0/3.2	&0.5/1.6/2.6\\
T=20 + Margin	&3.5/10.1/\textbf{11.8}	&3.4/12.0/14.9	&0.3/1.2/\textbf{1.4} &0.6/1.2/1.9	&0.5/1.9/2.9	&0.5/1.6/2.7\\
T=20 + Angle	&3.2/9.7/10.1	&3.9/11.9/13.3	&0.3/1.0/1.2 &0.6/1.6/2.0	&0.6/2.0/\textbf{3.5}	&0.5/1.7/\textbf{2.9}\\
Margin + Angle  &3.3/9.8/11.1	& 3.5/12.6/14.6	& 0.3/1.2/\textbf{1.4}  & 0.6/1.4/2.0	& 0.6/1.7/3.1	& 0.5/1.5/2.6\\\hline
\end{tabular}
\end{table*}

\subsection{Targeted Transfer of Ensemble Attacks}
\begin{figure*}
  \centering
  \begin{subfigure}{0.24\textwidth}
     \includegraphics[width=\linewidth]{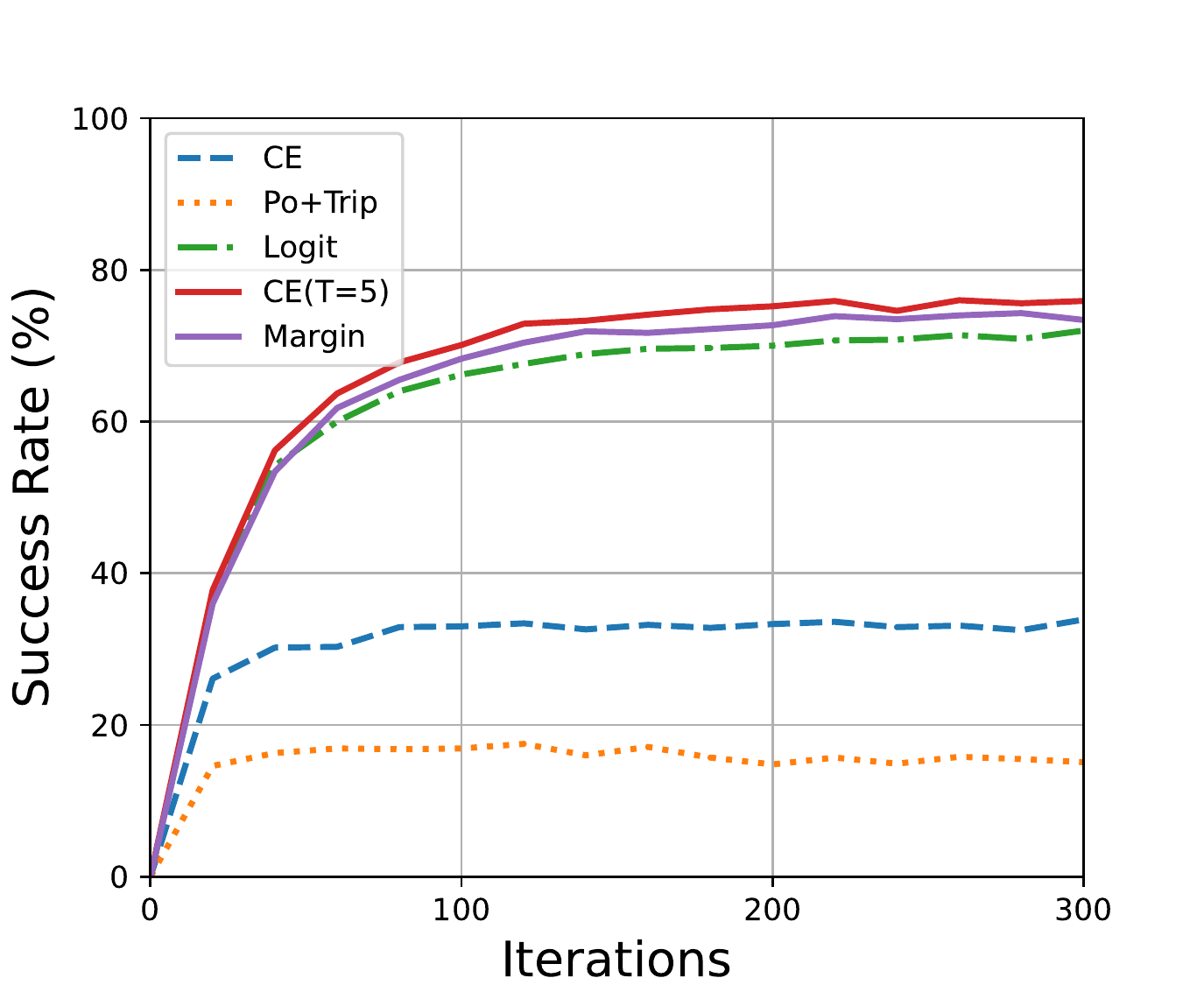}
    \caption{ResNet-50}
  \end{subfigure}
  \begin{subfigure}{0.24\textwidth}
     \includegraphics[width=\linewidth]{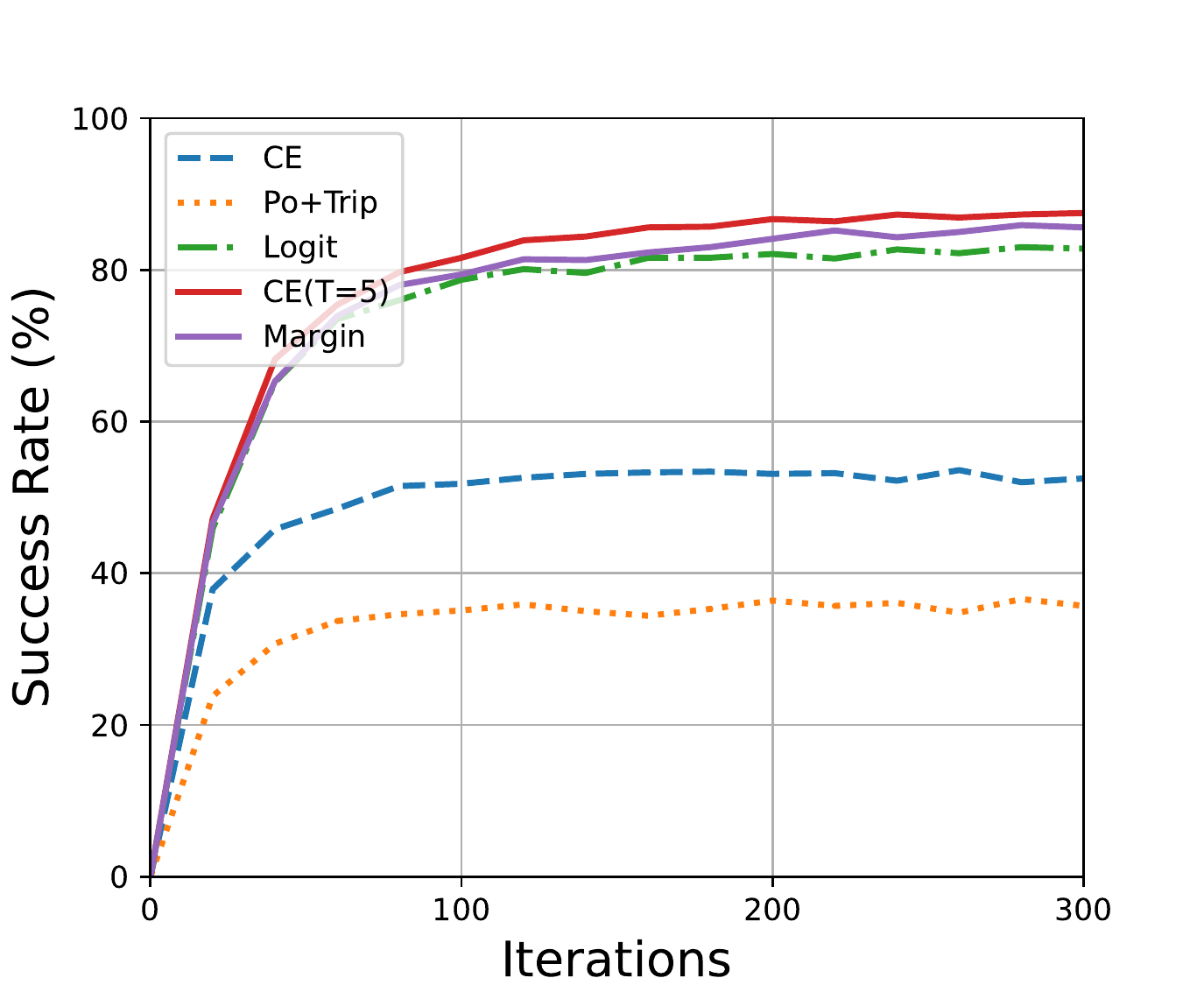}
    \caption{DenseNet-121}
  \end{subfigure}
  \begin{subfigure}{0.24\textwidth}
     \includegraphics[width=\linewidth]{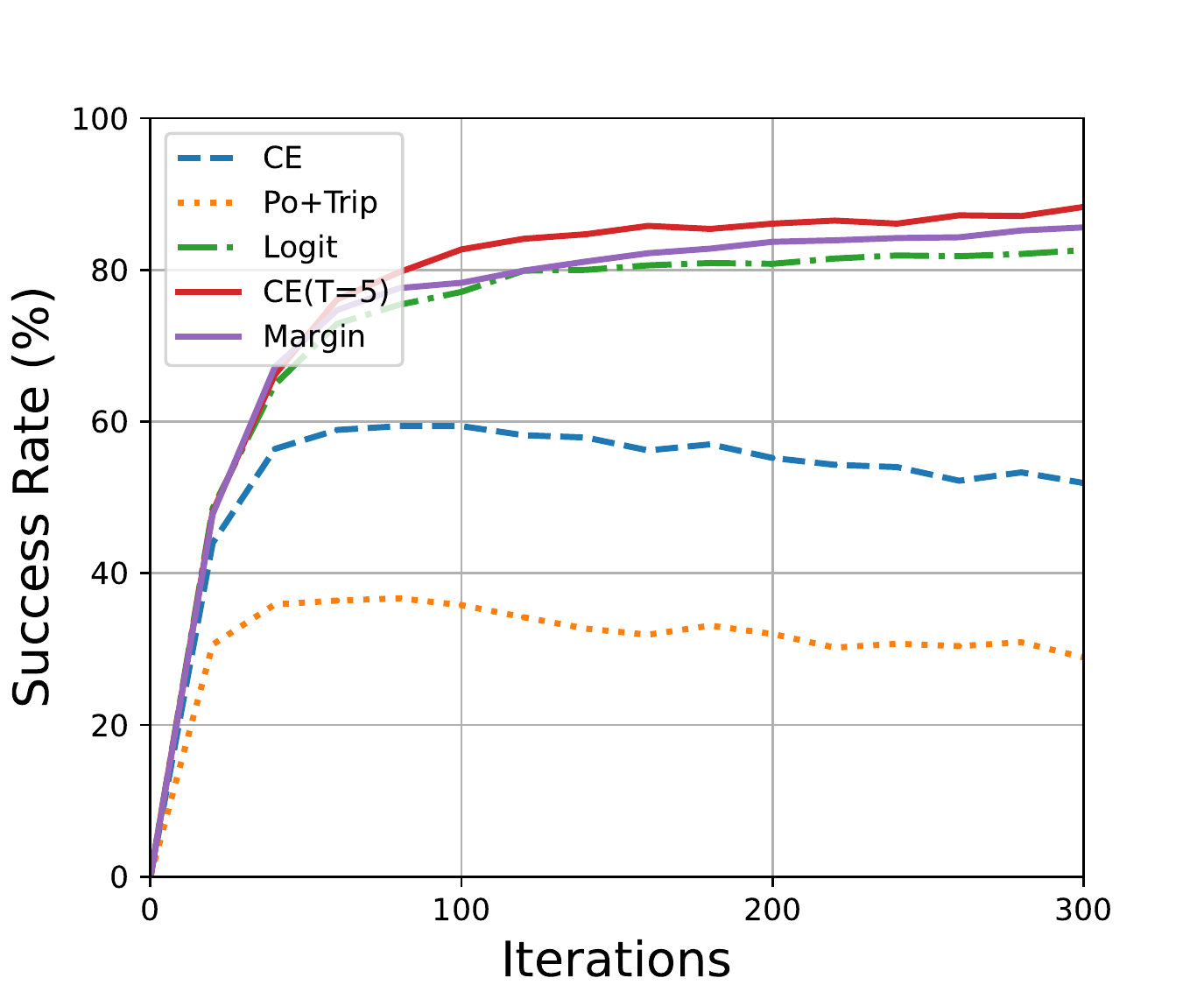}
    \caption{VGG-16}
  \end{subfigure}
  \begin{subfigure}{0.24\textwidth}
     \includegraphics[width=\linewidth]{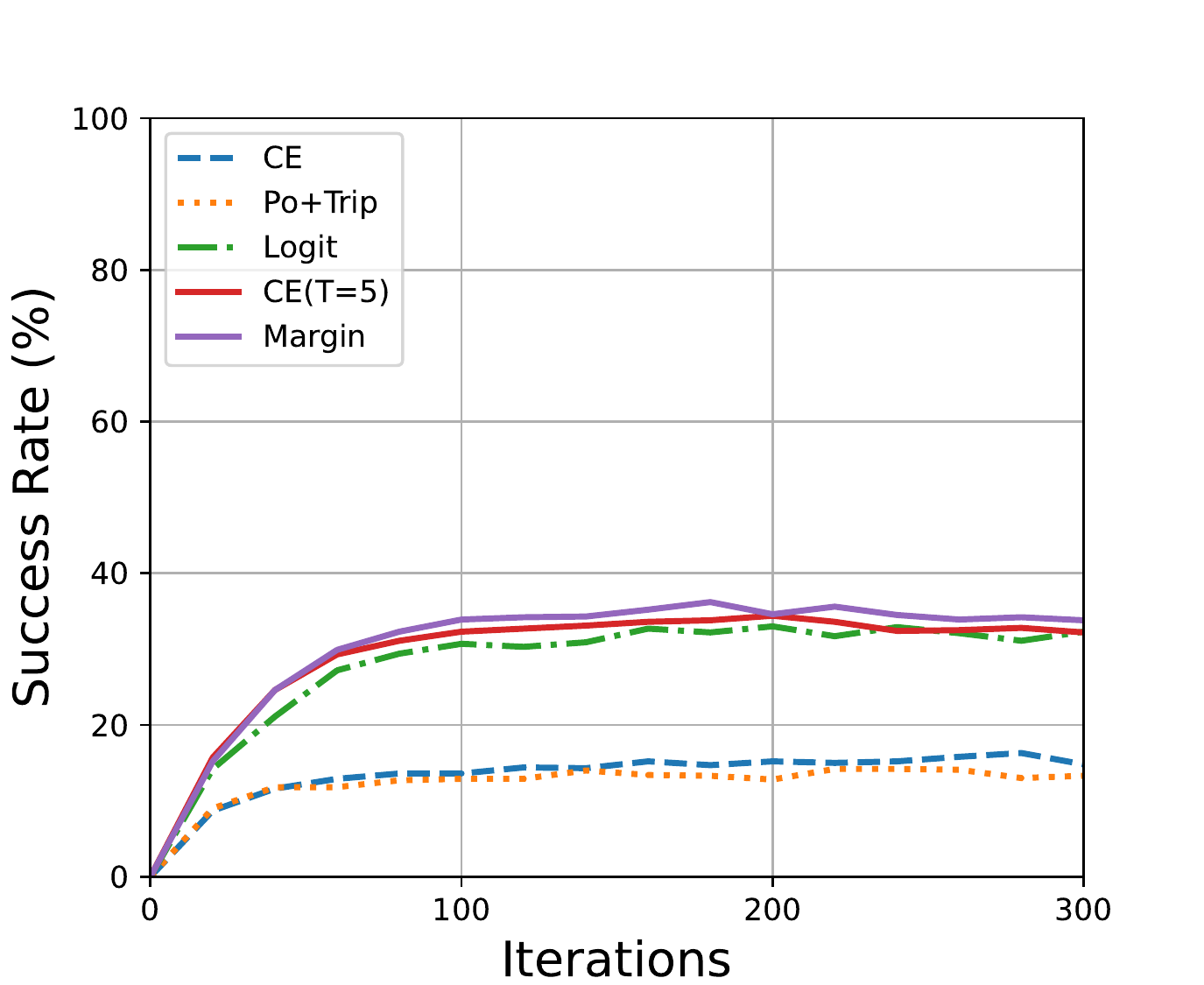}
    \caption{Inc-v3}
  \end{subfigure}
    \caption{Targeted transfer success rates (\%) in a realistic ensemble setting, where each hold-out target model shares no similar architecture with the source models used for the ensemble. We select four models---ResNet50, DenseNet121, VGG16 and Inc-v3. In each sub-figure, the caption indicates the name of the hold-out model and the adversarial examples are generated for the ensemble of the other three models.}
    \label{fig:ensemble}
\end{figure*}
Since ensemble attacks can significantly improve transferability, we also use an ensemble of white-box source models to optimize the perturbations for further evaluation.
Following the realistic ensemble setting in the {Logit}~\cite{zhao2021success}, we also select the four models---ResNet50, DenseNet121, VGG16, and Inc-v3, where each hold-out target model shares no similar architecture with the other three source models used for the ensemble. Besides, we simply assign equal weights to the source models during the training. 

The results are shown in Figure~\ref{fig:ensemble}, and we can obtain the following observations.
\textbf{1)} The Po+Trip produces the worst performance than the original CE and the Logit in the realistic ensemble setting. Besides, the original CE is also weak for ensemble transfer tasks.
\textbf{2)} After performing the logit calibration (CE~(T=5) and Margin), we can find a significant performance is boosted compared with the original CE. 
\textbf{3)} Meanwhile, our methods can outperform the Logit~\cite{zhao2021success}, comparing all results.  These results verify the effectiveness of logit calibration in the ensemble targeted attacks.

\subsection{Transfer Success Rates with Varied Targets}

Following the settings in~\cite{zhao2021success}, we further evaluate the performance of different calibration methods in the worse-case transfer scenario by gradually varying the target class from the highest-ranked to the lowest one. The ResNet-50 and the DesNet-121 models are used for training and evaluation, respectively.
From Table~\ref{tab:vary-targeted}, we can have the following findings: \textbf{1)} The three types of logit calibration methods can improve the targeted transfer success rate over the original CE. The angle-based calibration can achieve the best performance. \textbf{2)} The Temperature-based (T=5/10) and the Angle-based calibrations can outperform the Logit loss by a large margin, especially the Angle-based calibration. \textbf{3)} The margin-based calibration doesn't work well in this setting. When the target class is after the 200th, the performance of margin-based will be significantly lower than the Temperature-based and Angle-based calibrations.

\begin{table}[t]
\centering
\caption{Targeted transfer success rate (\%) when varying the target from the high-ranked class to low.}
\label{tab:vary-targeted}
\resizebox{\linewidth}{!}{
\begin{tabular}{l|cccccc} \hline
        &2nd    &10th	&200th	&500th	&800th	&1000th \\\hline
Logit	&83.7	&83.2	&74.5	&71.5	&64.9	&52.4 \\
CE	    &77.4	&58.6	&26.9	&23.7	&16.7	&7.0  \\     
CE/5	&91.3	&88.7	&77.1	&75.8	&70.1	&58.8 \\
CE/10	&89.0	&87.8	&81.0	&79.2	&73.5	&62.5 \\
Margin	&87.4	&81.7	&61.3	&51.6	&43.1	&23.0 \\
Angle	&92.4	&89.1	&80.3	&79.2	&76.1	&66.3 \\\hline
\end{tabular}
}
\end{table}

\subsection{Transfer-based Attacks on Google Cloud Vision}
Following the evaluation protocol in~\cite{zhao2021success}, we randomly select 100 images to conduct a real-world adversarial attack on the Google Cloud Vision API. The attacking performance is computed based on transfer-based attacks of the ensemble of four CNNs (\textit{i.e.}, Inc-V3, ResNet-50, Dense-121 and VGG-16). The results are shown in Table~\ref{tab:api}. We can find that the results of the Logit and CE (T=5) are very similar. But the Margin-based calibration performs worse than Logit and CE (T=5). These results reveal that our logit calibration-based targeted transfer attacks can cause a potential threat to the real-world Google Cloud Vision API.
\begin{table}[!ht]
    \centering
    \caption{Non-targeted and targeted transfer success rates (\%) on Google Cloud Vision API.}
    \label{tab:api}
    \begin{tabular}{l|ccc} \hline
                &  Logit    &CE (T=5)	& Margin\\ \hline
    Targeted	& 16	    & 15	    & 12    \\
Non-targeted	& 51	    & 53	    & 42    \\\hline
    \end{tabular}
\end{table}

\begin{table*}[t]
\centering
\caption{\textbf{Comparison with TTP~\cite{naseer2021generating} on Target Transferablity.} The averaged Top-1 targeted accuracy (\%) across 10 targets is computed with 49.95K ImageNet validation samples. Perturbation budget: $l_\infty  \leq 16$. * indicates the training surrogate model.}
\label{tab:ttp}
\begin{tabular}{c|c|ccccc|c} \hline
Dataset    &  Loss  & ResNet50* & VGG19$_{BN}$ & Dense121  & ResNet152 & WRN-50-2 & Average\\ \hline
\multirow{3}{*}{ImageNet} & TTP   & 97.02*   & 78.15   & 81.64        & 80.56     & 78.25  &83.12   \\ 
& CE       & 97.15* & 70.44   & 78.96        & 76.22     & 78.24  & 80.20 \\ 
& CE (T=5)     & \textbf{99.18*} & \textbf{86.65}   & \textbf{90.55}        & \textbf{90.30}     & \textbf{93.22}  & \textbf{91.98} \\ \hline

\multirow{2}{*}{Painting} & TTP     & 96.63* & 73.09   & 84.76        & 76.27     & 75.92  & 81.33 \\ 
& CE (T=5)    & \textbf{98.95*} & \textbf{82.97}   & \textbf{87.07}    & \textbf{87.81}  & \textbf{91.70} & \textbf{89.70}    \\ \hline
\end{tabular}
\end{table*}

\begin{table*}[t]
\caption{\textbf{Comparison with the ODI-MI-TI~\cite{byun2022improving} on Target Transferablity}. Targeted attack success rates (\%) against nine black-box target models with the four source models. For each attack, we also
report the average attack success rate (Avg.) against all black-box models.}
\centering
\label{tab:odi}
\resizebox{0.99\linewidth}{!}{
\begin{tabular}{lcccccccccc}
\toprule[1.2pt]
\textbf{Source:RN-50} & \multicolumn{9}{c}{Target model} & \multicolumn{1}{l}{} \\
 \cline{2-10}
Attack & VGG-16 & RN-18 & DN-121 & Inc-v3 & Inc-v4 & Mob-v2 & IR-v2 & Adv-Inc-v3 & Ens-adv-IR-v2 & Avg. \\
\hline
ODI-MI-TI & 76.8 & 77.0 & 86.8 & 67.4 & 55.4 & 66.8 & 48.0 & 0.7 & 1.7 & 53.4\\

CE~(T=5) & 80.7 & 80.7 & 88.9 & 68.1 & 56.7 & 71.0 & 51.8 & 0.6 & 1.2 & 55.5\\

Margin & 83.7 & 84.4 & 93.3 & 69.4 & 58.7 & 71.3 & 51.7 & 0.7 & 1.2 & 57.2\\
\toprule[1.2pt]

\textbf{Source:VGG-16} & \multicolumn{9}{c}{Target model} & \multicolumn{1}{l}{} \\
 \cline{2-10}
Attack & VGG-16 & RN-18 & DN-121 & Inc-v3 & Inc-v4 & Mob-v2 & IR-v2 & Adv-Inc-v3 & Ens-adv-IR-v2 & Avg. \\
\hline
ODI-MI-TI & 60.8 & 64.3 & 71.1 & 37.0 & 38.0 & 47.0 & 21.1 & 0.0 & 0.0 & 37.7\\

CE~(T=5) & 69.4 & 70.0 & 79.5 & 39.6 & 42.6 & 52.6 & 23.3 & 0.0 & 0.0 & 41.9 \\

Margin & 67.6 & 69.8 & 78.3 & 41.3 & 42.2 & 52.8 & 23.2 & 0.0 & 0.0 & 41.7\\
\toprule[1.2pt]

\textbf{Source:DN-121} & \multicolumn{9}{c}{Target model} & \multicolumn{1}{l}{} \\
 \cline{2-10}
Attack & VGG-16 & RN-18 & DN-121 & Inc-v3 & Inc-v4 & Mob-v2 & IR-v2 & Adv-Inc-v3 & Ens-adv-IR-v2 & Avg. \\
\hline
ODI-MI-TI & 64.5 & 63.4 & 71.6 & 53.5 & 46.4 & 44.2 & 38.3 & 0.4 & 0.7 & 42.6\\

CE(T=5) & 66.8 & 67.7 & 74.7 & 56.6 & 47.8 & 47.7 & 42.1 & 0.5 & 1.3  & 45.0 \\

Margin & 69.7 & 68.5 & 77.1 & 56.6 & 47.8 & 49.2 & 40.5 & 0.3 & 0.7 & 45.6 \\
\toprule[1.2pt]
\textbf{Source:Inc-v3} & \multicolumn{9}{c}{Target model} & \multicolumn{1}{l}{} \\
 \cline{2-10}
Attack & VGG-16 & RN-18 & DN-121 & Inc-v3 & Inc-v4 & Mob-v2 & IR-v2 & Adv-Inc-v3 & Ens-adv-IR-v2 & Avg. \\
\hline
ODI-MI-TI & 15.7 & 14.7 & 17.4 & 30.4 & 32.1 & 14.1 & 26.9 & 0.3 & 0.6& 16.9 \\

CE(T=5) & 15.0 & 16.2 & 18.7 & 34.5 & 31.5 & 15.3 & 27.8 & 0.2 & 0.5 & 17.7  \\

Margin & 13.5 & 15.3 & 16.6 & 31.4 & 31.3 & 13.2 & 24.9 & 0.3 & 0.3 & 16.3 \\
\toprule[1.2pt]
\end{tabular}}
\end{table*}
\subsection{Evaluating Logits Calibration in TTP and ODI}
In this section, we further evaluate the proposed temperature-based logit calibration in the GAN-based targeted attacks (TTP~\cite{naseer2021generating}) and the most recently published object-based diverse input (ODI) method~\cite{byun2022improving}.

Following the setting in TTP~\cite{naseer2021generating}, we sample 50K images from the ImageNet training set and 50K images from the Painting dataset\footnote{\url{https://www.kaggle.com/c/painter-by-numbers}}, which are used to train the targeted generators from different source domains. Instead of using the distribution matching and neighborhood similarity matching loss~\cite{naseer2021generating}, we only use the cross-entropy function for training the targeted generators while keeping other settings identical. More training and evaluation details used by TTP can be referred to ~\cite{naseer2021generating}. We use the ResNet50 as the surrogate model and report the results in Table~\ref{tab:ttp}. We have the following findings from Table~\ref{tab:ttp},. \textbf{1)} By using ImageNet as the training dataset, the TTP shows better transferability than the CE in attacking other black-box models. The average targeted accuracy of TTP is around 3\% higher than that of CE. \textbf{2)} After downscaling the logit by 5 in the CE loss function (T=5), we can observe a significant boost of the Top-1 targeted accuracy for all models, reaching an average value of 91.98\% (ImageNet). \textbf{3)} For both ImageNet and Painting as the training source, the CE (T=5) can surpass the TTP by a large margin (91.98\% vs. 83.12\% \& 89.70\% vs. 81.33\%). Note that, compared to TTP, our logit calibration has the benefit without any data from the target class. These experimental results demonstrate that the proposed temperate-based logit calibration is also effective in training generator-based targeted attackers. 

For the ODI~\cite{byun2022improving}, we follow the setting in ODI, which improves the targeted transferability by projecting the adversarial examples on 3D objects and uses the Logit loss~\cite{zhao2021success} for optimizing the targeted adversarial examples.  To compare with the original ODI, we use the "Package, Pillow, Book" as the 3D objects and  simply replace the Logit loss~\cite{zhao2021success} with the CE (T=5) and Margin-based calibration for training with 300 iterations. Besides, we also select the same six additional models for more comprehensive comparisons, including ResNet18 (RN-18), a lightweight model MobileNet-v2
(Mob-v2)~\cite{sandler2018mobilenetv2}, Inception ResNet-v2 (IR-v2)~\cite{szegedy2017inception},  Inception-v4 (Inc-v4)~\cite{szegedy2017inception} and two adversarial trained models (adversarially trained Inc-v3 (Adv-Inc-v3)~\cite{kurakin2018adversarial}, and ensemble-adversarially trained IR-v2 (Ens-adv-IR-v2)~\cite{kurakin2018adversarial}). The results are reported in Table~\ref{tab:odi}. We have the following observations from Table~\ref{tab:odi}: \textbf{1)} For the source model ResNet-50, VGG-16, and Dense-121, the CE (T=5) and Margin-based calibration significantly outperform the Logit loss when using the ODI~\cite{byun2022improving} for data augmentation. Besides, the Margin-based calibration can perform slightly better than the CE (T=5). \textbf{2)} For the Inception-V3, there is no significant difference between the Logit, CE~(T=5) and Margin-based calibration.

\section{Conclusion}
In this study, we analyze the logit margin in different loss functions for the transferable targeted attack and find that the margin will quickly get saturated in the CE loss and thus limits the transferability. To deal with this issue, we propose to use logit calibrations in the CE loss function, including Temperature-based, Margin-based, and Angle-based. Experimental results verify the effectiveness of using the logit calibration in the CE loss function for crafting transferable targeted adversarial samples. The proposed logit calibrations are simple and easy to implement, which can achieve state-of-the-art performance in transferable targeted attacks. 

\section*{Potential Social Impact}


Our findings in targeted transfer attacks can potentially motivate the AI community to design more robust defenses against transferable attacks. In the long run, it may be directly used for suitable social applications, such as protecting privacy. 
Contrariwise, some applications may use targeted transferable attacks in a harmful manner to damage the outcome of AI systems, especially in speech recognition and facial verification. Finally, we firmly believe that our investigation in this study can provide valuable insight for future researchers of using logit calibration for both adversarial attack and defense.
\bibliographystyle{IEEEtran}
\bibliography{ref}

\begin{thebibliography}{10}
\providecommand{\url}[1]{#1}
\csname url@samestyle\endcsname
\providecommand{\newblock}{\relax}
\providecommand{\bibinfo}[2]{#2}
\providecommand{\BIBentrySTDinterwordspacing}{\spaceskip=0pt\relax}
\providecommand{\BIBentryALTinterwordstretchfactor}{4}
\providecommand{\BIBentryALTinterwordspacing}{\spaceskip=\fontdimen2\font plus
\BIBentryALTinterwordstretchfactor\fontdimen3\font minus
  \fontdimen4\font\relax}
\providecommand{\BIBforeignlanguage}[2]{{%
\expandafter\ifx\csname l@#1\endcsname\relax
\typeout{** WARNING: IEEEtran.bst: No hyphenation pattern has been}%
\typeout{** loaded for the language `#1'. Using the pattern for}%
\typeout{** the default language instead.}%
\else
\language=\csname l@#1\endcsname
\fi
#2}}
\providecommand{\BIBdecl}{\relax}
\BIBdecl

\bibitem{simonyan2014very}
K.~Simonyan and A.~Zisserman, ``Very deep convolutional networks for
  large-scale image recognition,'' in \emph{ICLR}, 2015.

\bibitem{long2015fully}
J.~Long, E.~Shelhamer, and T.~Darrell, ``Fully convolutional networks for
  semantic segmentation,'' in \emph{CVPR}, 2015.

\bibitem{ren2015faster}
S.~Ren, K.~He, R.~Girshick, and J.~Sun, ``Faster r-cnn: Towards real-time
  object detection with region proposal networks,'' \emph{NeurIPS}, 2015.

\bibitem{goodfellow2014explaining}
I.~J. Goodfellow, J.~Shlens, and C.~Szegedy, ``Explaining and harnessing
  adversarial examples,'' in \emph{ICLR}, 2015.

\bibitem{wang2020towards}
Q.~Wang, B.~Zheng, Q.~Li, C.~Shen, and Z.~Ba, ``Towards query-efficient
  adversarial attacks against automatic speech recognition systems,''
  \emph{IEEE Transactions on Information Forensics and Security}, vol.~16, pp.
  896--908, 2020.

\bibitem{wu2020audio}
J.~Wu, B.~Chen, W.~Luo, and Y.~Fang, ``Audio steganography based on iterative
  adversarial attacks against convolutional neural networks,'' \emph{IEEE
  transactions on information forensics and security}, vol.~15, pp. 2282--2294,
  2020.

\bibitem{zhong2020towards}
Y.~Zhong and W.~Deng, ``Towards transferable adversarial attack against deep
  face recognition,'' \emph{IEEE Transactions on Information Forensics and
  Security}, vol.~16, pp. 1452--1466, 2020.

\bibitem{dong2019efficient}
Y.~Dong, H.~Su, B.~Wu, Z.~Li, W.~Liu, T.~Zhang, and J.~Zhu, ``Efficient
  decision-based black-box adversarial attacks on face recognition,'' in
  \emph{Proceedings of the IEEE/CVF Conference on Computer Vision and Pattern
  Recognition}, 2019, pp. 7714--7722.

\bibitem{ding2021beyond}
W.~Ding, X.~Wei, R.~Ji, X.~Hong, Q.~Tian, and Y.~Gong, ``Beyond universal
  person re-identification attack,'' \emph{IEEE transactions on information
  forensics and security}, vol.~16, pp. 3442--3455, 2021.

\bibitem{yang2022towards}
F.~Yang, J.~Weng, Z.~Zhong, H.~Liu, Z.~Wang, Z.~Luo, D.~Cao, S.~Li, S.~Satoh,
  and N.~Sebe, ``Towards robust person re-identification by defending against
  universal attackers,'' \emph{IEEE Transactions on Pattern Analysis and
  Machine Intelligence}, 2022.

\bibitem{dong2018boosting}
Y.~Dong, F.~Liao, T.~Pang, H.~Su, J.~Zhu, X.~Hu, and J.~Li, ``Boosting
  adversarial attacks with momentum,'' in \emph{CVPR}, 2018.

\bibitem{dong2019evading}
Y.~Dong, T.~Pang, H.~Su, and J.~Zhu, ``Evading defenses to transferable
  adversarial examples by translation-invariant attacks,'' in \emph{CVPR},
  2019.

\bibitem{cohen2019certified}
J.~Cohen, E.~Rosenfeld, and Z.~Kolter, ``Certified adversarial robustness via
  randomized smoothing,'' in \emph{ICML}, 2019.

\bibitem{tramer2017ensemble}
F.~Tram{\`e}r, A.~Kurakin, N.~Papernot, I.~Goodfellow, D.~Boneh, and
  P.~McDaniel, ``Ensemble adversarial training: Attacks and defenses,'' in
  \emph{ICLR}, 2018.

\bibitem{xie2019improving}
C.~Xie, Z.~Zhang, Y.~Zhou, S.~Bai, J.~Wang, Z.~Ren, and A.~L. Yuille,
  ``Improving transferability of adversarial examples with input diversity,''
  in \emph{CVPR}, 2019.

\bibitem{byun2022improving}
J.~Byun, S.~Cho, M.-J. Kwon, H.-S. Kim, and C.~Kim, ``Improving the
  transferability of targeted adversarial examples through object-based diverse
  input,'' in \emph{CVPR}, 2022.

\bibitem{li2022decision}
X.-C. Li, X.-Y. Zhang, F.~Yin, and C.-L. Liu, ``Decision-based adversarial
  attack with frequency mixup,'' \emph{IEEE Transactions on Information
  Forensics and Security}, vol.~17, pp. 1038--1052, 2022.

\bibitem{liu2016delving}
Y.~Liu, X.~Chen, C.~Liu, and D.~Song, ``Delving into transferable adversarial
  examples and black-box attacks,'' in \emph{ICLR}, 2016.

\bibitem{lin2019nesterov}
J.~Lin, C.~Song, K.~He, L.~Wang, and J.~E. Hopcroft, ``Nesterov accelerated
  gradient and scale invariance for adversarial attacks,'' in \emph{ICLR},
  2020.

\bibitem{huang2019enhancing}
Q.~Huang, I.~Katsman, H.~He, Z.~Gu, S.~Belongie, and S.-N. Lim, ``Enhancing
  adversarial example transferability with an intermediate level attack,'' in
  \emph{CVPR}, 2019.

\bibitem{wu2020skip}
D.~Wu, Y.~Wang, S.-T. Xia, J.~Bailey, and X.~Ma, ``Skip connections matter: On
  the transferability of adversarial examples generated with resnets,'' in
  \emph{ICLR}, 2020.

\bibitem{guo2020backpropagating}
Y.~Guo, Q.~Li, and H.~Chen, ``Backpropagating linearly improves transferability
  of adversarial examples,'' in \emph{NeurIPS}, 2020.

\bibitem{xu2022bounded}
Q.~Xu, G.~Tao, and X.~Zhang, ``Bounded adversarial attack on deep content
  features,'' in \emph{CVPR}, 2022.

\bibitem{xiong2022stochastic}
Y.~Xiong, J.~Lin, M.~Zhang, J.~E. Hopcroft, and K.~He, ``Stochastic variance
  reduced ensemble adversarial attack for boosting the adversarial
  transferability,'' in \emph{CVPR}, 2022.

\bibitem{li2020towards}
M.~Li, C.~Deng, T.~Li, J.~Yan, X.~Gao, and H.~Huang, ``Towards transferable
  targeted attack,'' in \emph{CVPR}, 2020.

\bibitem{zhao2021success}
Z.~Zhao, Z.~Liu, and M.~Larson, ``On success and simplicity: A second look at
  transferable targeted attacks,'' \emph{NeurIPS}, vol.~34, 2021.

\bibitem{hinton2015distilling}
G.~Hinton, O.~Vinyals, and J.~Dean, ``Distilling the knowledge in a neural
  network (2015),'' \emph{arXiv preprint arXiv:1503.02531}, 2015.

\bibitem{szegedy2015going}
C.~Szegedy, W.~Liu, Y.~Jia, P.~Sermanet, S.~Reed, D.~Anguelov, D.~Erhan,
  V.~Vanhoucke, and A.~Rabinovich, ``Going deeper with convolutions,'' in
  \emph{CVPR}, 2015.

\bibitem{kurakin2018adversarial}
A.~Kurakin, I.~J. Goodfellow, and S.~Bengio, ``Adversarial examples in the
  physical world,'' in \emph{Artificial intelligence safety and security},
  2018, pp. 99--112.

\bibitem{naseer2021generating}
M.~Naseer, S.~Khan, M.~Hayat, F.~S. Khan, and F.~Porikli, ``On generating
  transferable targeted perturbations,'' in \emph{ICCV}, 2021.

\bibitem{kurakin2016adversarial}
A.~Kurakin, I.~Goodfellow, and S.~Bengio, ``Adversarial machine learning at
  scale,'' \emph{arXiv preprint arXiv:1611.01236}, 2016.

\bibitem{inkawhich2020transferable}
N.~Inkawhich, K.~J. Liang, L.~Carin, and Y.~Chen, ``Transferable perturbations
  of deep feature distributions,'' \emph{arXiv preprint arXiv:2004.12519},
  2020.

\bibitem{inkawhich2020perturbing}
N.~Inkawhich, K.~Liang, B.~Wang, M.~Inkawhich, L.~Carin, and Y.~Chen,
  ``Perturbing across the feature hierarchy to improve standard and strict
  blackbox attack transferability,'' \emph{NeruIPS}, 2020.

\bibitem{poursaeed2018generative}
O.~Poursaeed, I.~Katsman, B.~Gao, and S.~Belongie, ``Generative adversarial
  perturbations,'' in \emph{CVPR}, 2018.

\bibitem{naseer2019cross}
M.~M. Naseer, S.~H. Khan, M.~H. Khan, F.~Shahbaz~Khan, and F.~Porikli,
  ``Cross-domain transferability of adversarial perturbations,''
  \emph{NeurIPS}, 2019.

\bibitem{liu2017sphereface}
W.~Liu, Y.~Wen, Z.~Yu, M.~Li, B.~Raj, and L.~Song, ``Sphereface: Deep
  hypersphere embedding for face recognition,'' in \emph{CVPR}, 2017.

\bibitem{deng2019arcface}
J.~Deng, J.~Guo, N.~Xue, and S.~Zafeiriou, ``Arcface: Additive angular margin
  loss for deep face recognition,'' in \emph{CVPR}, 2019.

\bibitem{deng2009imagenet}
J.~Deng, W.~Dong, R.~Socher, L.-J. Li, K.~Li, and L.~Fei-Fei, ``Imagenet: A
  large-scale hierarchical image database,'' in \emph{CVPR}.\hskip 1em plus
  0.5em minus 0.4em\relax Ieee, 2009, pp. 248--255.

\bibitem{he2016deep}
K.~He, X.~Zhang, S.~Ren, and J.~Sun, ``Deep residual learning for image
  recognition,'' in \emph{CVPR}, 2016.

\bibitem{huang2017densely}
G.~Huang, Z.~Liu, L.~Van Der~Maaten, and K.~Q. Weinberger, ``Densely connected
  convolutional networks,'' in \emph{CVPR}, 2017.

\bibitem{szegedy2016rethinking}
C.~Szegedy, V.~Vanhoucke, S.~Ioffe, J.~Shlens, and Z.~Wojna, ``Rethinking the
  inception architecture for computer vision,'' in \emph{CVPR}, 2016.

\bibitem{sandler2018mobilenetv2}
M.~Sandler, A.~Howard, M.~Zhu, A.~Zhmoginov, and L.-C. Chen, ``Mobilenetv2:
  Inverted residuals and linear bottlenecks,'' in \emph{Proceedings of the IEEE
  conference on computer vision and pattern recognition}, 2018, pp. 4510--4520.

\bibitem{szegedy2017inception}
C.~Szegedy, S.~Ioffe, V.~Vanhoucke, and A.~Alemi, ``Inception-v4,
  inception-resnet and the impact of residual connections on learning,'' in
  \emph{Proceedings of the AAAI conference on artificial intelligence},
  vol.~31, no.~1, 2017.

\end{thebibliography}
\end{document}